\pdfoutput=1
\documentclass{article}

\usepackage[preprint]{neurips_2024}

\usepackage[utf8]{inputenc} % allow utf-8 input
\usepackage{graphicx}
\usepackage[T1]{fontenc}    % use 8-bit T1 fonts
\usepackage{hyperref}       % hyperlinks
\usepackage{url}            % simple URL typesetting
\usepackage{booktabs}       % professional-quality tables
\usepackage{amsfonts}       % blackboard math symbols
\usepackage{nicefrac}       % compact symbols for 1/2, etc.
\usepackage{microtype}      % microtypography
\usepackage{xcolor}         % colors
\usepackage{multirow}
\usepackage{wrapfig}
\usepackage{amsmath}
\usepackage{gensymb}
\usepackage{enumitem}
\usepackage{arydshln}

\title{Structure-Aware Gaussians through Lightweight Information Shaping}

\author{%
  Yunchao Zhang\textsuperscript{1},
  Guandao Yang\textsuperscript{2},
  Leonidas Guibas\textsuperscript{2},
  Yanchao Yang\textsuperscript{1}\\
  \textsuperscript{1} The University of Hong Kong \ \ \ \ \ \textsuperscript{2} Stanford University\\
}

\begin{document}

\maketitle

\vspace{-0.5cm}
\begin{abstract}

3D Gaussians, 
as an explicit scene representation, 
typically involve 
thousands to millions of elements per scene. 
This makes it 
challenging to control 
the scene in ways 
that reflect the underlying semantics, 
where the number of independent entities 
is typically much smaller. 
Especially, 
if one wants to animate or edit objects in the scene, 
as this requires coordination among the many Gaussians
involved in representing each object. 
To address this issue, 
we develop a mutual information shaping technique 
that enforces resonance and coordination
between correlated Gaussians 
via a Gaussian attribute decoding network. 
Such correlations can be learned 
from putative 2D object masks in different views. 
By approximating the 
mutual information with 
the gradients concerning the network parameters, 
our method ensures consistency 
between scene elements 
and enables efficient scene editing 
by operating on network parameters rather than massive Gaussians.
In particular, 
we develop an effective contrastive learning pipeline
with lightweight optimization to shape the attribute decoding network,
while ensuring that the shaping (consistency) is maintained during continuous edits, 
avoiding re-shaping after parameter changes. 
Notably, 
our training only touches 
a small fraction of all Gaussians in the scene 
yet attains the desired correlated behavior 
according to the underlying scene structure. 
The proposed technique 
is evaluated on challenging scenes and demonstrates significant performance improvements 
in 3D object segmentation and promoting scene interactions, 
while inducing 
low computation and memory requirements. 
Our code and trained models 
will be made available.

% We propose addressing the issue of inconsistent dynamics in neural scene representations such that objects in the scene can be perturbed to synthesize coherent movements without falling apart. To resolve this issue, we develop a mutual information shaping technique that enforces correlation between different Gaussians in dynamic 3D Gaussian Splatting models. By approximating mutual information between the Jacobians of the motion, the method ensures consistent motion within the network under various perturbations. Particularly, an efficuent training pipeline is developed to shape the motion network at a single time step, avoiding the need for re-shaping throughout the sequence. The proposed technique is evaluated on challenging scenes and demonstrates state-of-the-art performance in promoting consistent dynamics and 3D object segmentation while inducing minimal computation and memory requirements. Our code and trained models will be made available to facilitate future research.

%Structure-Aware?
%  \keywords{dynamic scene representation \and consistent motion}
\end{abstract}
\section{Introduction}

Open-world 3D scene understanding and editing 
play important roles in gaming, AR/VR, and robotics applications. 
Recently, 
3D Gaussian Splatting (3DGS) \citep{kerbl20233d}, 
has significantly advanced 
both the rendering quality and inference efficiency 
in 3D learning by representing scenes through a set of 3D Gaussians. 
Moreover, follow-up works 
in scene editing \citep{cen2023segment,zhou2023feature,ye2023gaussian} propose augmenting each Gaussian's parameters 
with task-relevant attributes, 
allowing users to edit scenes by selecting Gaussians based on these attributes and modifying their colors or positions.
Despite the progress 
with 3DGS-based scene editing, 
the use of point-wise 3D Gaussians 
%as a low-level scene representation optimized for novel-view synthesis 
presents limitations, 
among which, 
the fact that the number of distinct and interactive entities in a scene is typically far fewer than the number of Gaussians is neglected. 
Furthermore, 
augmenting attributes of Gaussians 
introduce significant storage overhead and do not fundamentally address the lack of correlation between Gaussians.
%While a sparse, unstructured cloud of independent Gaussians may be sufficient for tasks such as novel-view synthesis, this approach proves inadequate for structure-aware or interactive tasks, such as semantic queries and scene editing.

The lack of intrinsic correlation 
among scene elements 
renders a clear contrast between the editing of 3D Gaussians and the perturbation in the real world. 
For example, 
%editing operations in the physical world are intuitive and efficient. 
real-world scenes consist of particles 
that aggregate into cohesive objects 
through physical or chemical bonds, 
forming highly organized, interdependent structures. 
Interactions with a small part of an object can induce changes throughout the entire entity due to these intrinsic correlations. 
Such a phenomenon implies 
a high degree of mutual information 
between particles of the same entity, 
and even across distinct objects. 
Ideally, 
3D scene representations should capture these correlation structures, 
facilitating efficient learning 
and interaction for tasks like scene editing. 
If so, 
modifying the state of a single element 
could result in 
coherent adjustment in the states of all correlated elements, 
which promotes efficiency 
by eliminating the need to individually examine and edit each Gaussian.

% \yc{after we describe the general topic, we can describe the current state of the art, but I feel like we like to introduce the work that compresses the Gaussians, which is the most related domain? but it also depends on how we describe the method.}
% Editing tasks, by their nature, are more concerned with relationships rather than individual values.

\begin{figure}[!t]
    \includegraphics[width=\textwidth]{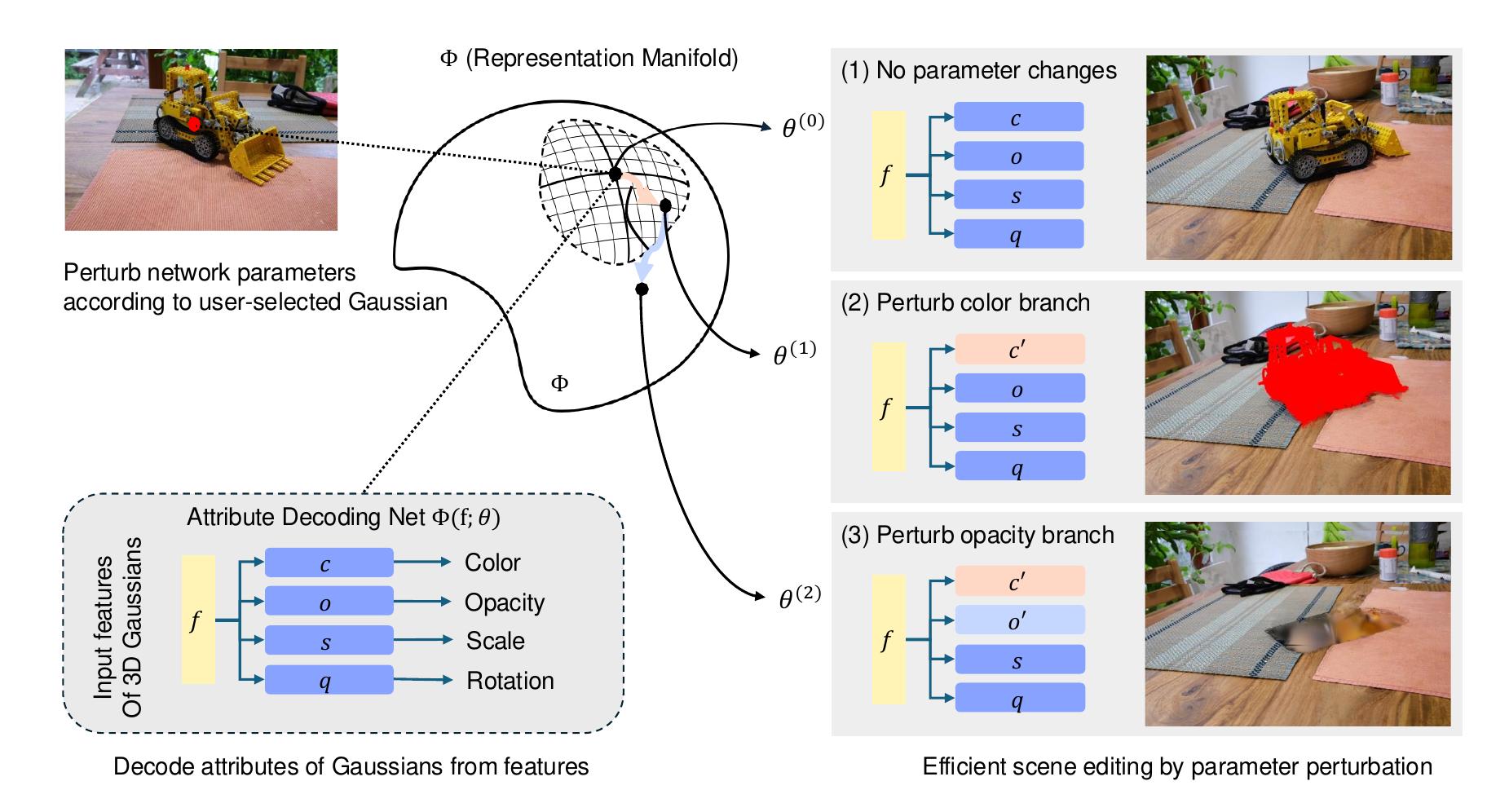}
    \caption{
    The proposed 
    mutual information shaping 
    of the attribute decoding network 
    based on 3D Gaussian Splatting~\cite{kerbl20233d} 
    can capture the underlying structure 
    of the scene, 
    while maintaining the correlations 
    after consecutive parameter changes according to user-selected Gaussian (specified in Sec.~\ref{sec:edit}).
    It promotes efficient scene editing
    by perturbing the network parameters, including re-colorization, segmentation, object removal, etc.
    % \yc{to revise: (1) get perturbation is confusing; color perturbation and opacity perturbation with arrows of different colors; dashed line points to arrows; make the figure more compact.}
    }
    \label{fig:teaser}
    \vspace{-2.5ex}
\end{figure}

To enable efficient object-level interactive tasks, 
we propose 
the key is to enforce the inherent correlation structure within the scene representation, 
with which, 
the propagation of changes across correlated Gaussians can be easily achieved. 
Instead of assigning additional attributes 
to each Gaussian for grouping purposes, 
we introduce a novel learning scheme 
that encodes the correlations 
into the representation 
and an effective approach for manipulating scene.
Note that, 
a few 3DGS-based compression methods have exploited 
similarities between Gaussians to reduce storage cost, 
such as clustering Gaussians with similar parameter values~\citep{Fan2023LightGaussianU3}, 
introducing anchors to group nearby Gaussians in different spatial regions~\citep{Lu2023ScaffoldGSS3}, 
or organizing 3D space with feature closeness~\citep{Chen2024HACHA}.
These techniques primarily target scene reconstruction and rendering, 
however, 
we propose an explicit modeling 
of the correlations 
from the perspective of mutual information 
maximization to enable efficient scene editing.

We develop upon the common practice 
of 3DGS compression methods, 
where Gaussian attributes (used for reconstruction) are decoded from their features, 
and we introduce a mutual information shaping 
scheme to explicitly enforce the correlations between Gaussians via the feature-decoding network.
This scheme ensures that 
perturbing the network parameters 
results in coherent changes among correlated Gaussians, 
enabling controllable and semantically meaningful interactions with objects in the rendered scene. 
Our strategy circumvents the need to estimate and encode pairwise feature correlations among millions of Gaussians. 
Instead, it constructs a well-organized tangent space that captures the underlying structure of the scene within the network.

Specifically, 
we devise a shaping loss 
that maximizes the mutual information (MI) 
between Gaussians belonging to the same entity (as evidenced by putative metrics),
which correlates Gaussians associated with the same object and de-correlates otherwise. 
Furthermore, 
mutual information between Gaussians 
is approximated by the cosine similarity between the Jacobians with respect to the perturbed weights. 
This legitimates the tangent space 
of the feature-decoding network parameters 
to encode the desired correlation structure. 
However, 
simply shaping the tangent space 
{\it does not} guarantee that the perturbed network will preserve this correlation structure when the shaped weights are modified, 
i.e., the after-modification parameters may not have a shaped tangent space, 
thus, making it incapable of consistent perturbations across multiple editing operations. 
Another challenge lies in the computational and memory overhead 
required to calculate Jacobians, 
as this necessitates storing the full computational graph regarding the derivative of the Jacobians --
a significant cost for 3DGS, 
where storage and training efficiency are critical concerns.

To further overcome these challenges, 
we propose shaping the activations of the perturbed weights instead of the Jacobians. 
We prove that this technique 
ensures that the network undergoes a conformal transformation in its tangent space, 
maintaining the correlation structure 
and enabling consistent editing across sequential operations. 
In other words, 
the proposed is a sufficient condition 
for mutual information maximization 
while inducing the benefits 
not possessed by directly shaping Jacobians.
Finally, 
we develop an efficient pipeline 
that shapes the feature-decoding network with consistency preserved along a sequence of perturbations, 
while being computationally efficient.
This pipeline is fully automated and light-weight.
%leveraging 2D masks generated by SAM~\cite{kirillov2023segany}. 
We evaluate the proposed MI shaping technique 
across various scene-editing applications, 
including 3D object removal, inpainting, colorization, and scene recomposition. 

In summary, 
1) we propose a novel approach 
that leverages semantic correlations 
between 3D Gaussians for efficient scene editing; 
2) We design an efficient training pipeline 
that shapes the scene representation once and 
ensures that the perturbed representation 
continues to support coherent edits 
without reshaping along the perturbation trajectory; 
3) We demonstrate that 
our MI shaping technique affects 
only approximately $\sim7\%$ of all Gaussians 
during training, 
while existing scene-editing methods require optimizing the entire set of Gaussians. 
Despite the lightweightness, 
our method achieves significant performance gains in tasks such as 3D segmentation, object removal, and inpainting, while substantially reducing computational and memory overhead.
\section{Related works}

% \subsection{Neural representations for 3D reconstruction} 
% In recent years, implicit neural representations have revolutionized 3D reconstruction, with NeRF~\citep{mildenhall2021nerf} standing out among works~\citep{barron2021mip, deng2022depth,karras2021alias,liu2020neural,martin2021nerf, zhang2023nerflets} that parameterize 3D scenes using neural networks. A series of subsequent works have extended the NeRF framework, proposing various strategies to compress large-scale 3D scenes~\citep{garbin2021fastnerf, muller2022instant,Reiser2021KiloNeRFSU} by incorporating multi-resolution hash tables or voxel grids. Drawing from explicit modeling, approaches such as~\citet{fridovich2022plenoxels,xu2022point,kerbl20233d} integrate traditional point clouds to represent scenes. Among these, 3D Gaussian Splatting (3DGS)\citep{kerbl20233d} reconstructs 3D scenes using 3D Gaussians for real-time rendering. Neural representations have also enabled a range of downstream tasks, including semantic understanding\citep{zhi2021place,Bao2023SINESI,Wang2023SemanticIE,Gao2022ReconstructingPS, Li2024SGSSLAMSG}, segmentation~\citep{fu2022panoptic,ye2023gaussian, cen2023segment}, and 3D content creation~\citep{tang2023dreamgaussian,Chen2023Textto3DUG}. However, these methods are largely driven by first-order supervision, overlooking potential higher-order correlations within the scene representation.

\paragraph{Neural representations for 3D reconstruction} 
Implicit neural representations 
have recently transformed the field of 3D reconstruction, 
such as Neural Radiance Fields (NeRF)~\citep{mildenhall2021nerf} 
and its extensions~\citep{barron2022mip,deng2022depth,karras2021alias,liu2020neural,martin2021nerf,zhang2023nerflets}, 
which utilize neural networks to parameterize 3D scenes, 
enabling detailed and accurate reconstructions.
Subsequent advancements have focused on optimizing the NeRF framework for large-scale scenes. 
Techniques such as multi-resolution hash tables and voxel grids have been proposed to compress scene representations~\citep{garbin2021fastnerf,muller2022instant,Reiser2021KiloNeRFSU}. 
Additionally, 
hybrid approaches have incorporated explicit modeling strategies, 
leveraging point clouds to enhance scene representation~\citep{fridovich2022plenoxels,xu2022point,kerbl20233d}. 
Notably, 
3D Gaussian Splatting (3DGS)~\citep{kerbl20233d} employs 
point-wise Gaussians to achieve efficient training and rendering.
These 3D representations 
are useful for various downstream tasks, such as semantic understanding~\citep{zhi2021place,Bao2023SINESI,Wang2023SemanticIE,Gao2022ReconstructingPS,Li2024SGSSLAMSG}, segmentation~\citep{fu2022panoptic,ye2023gaussian,cen2023segment}, 
and 3D content creation~\citep{tang2023dreamgaussian,Chen2023Textto3DUG}. 
Nevertheless, 
current methods predominantly rely on first-order supervision, 
often neglecting the higher-order correlations encoded in the scene representation.

% \subsection{Scene Representations for 3D Editing}
% Despite advancements in scene reconstruction, editing scenes with human interaction remains a significant challenge. Existing neural representation-based approaches to scene editing \citep{wang2022clip,Yuan2022NeRFEditingGE,schwarz2020graf,Wang2023ProteusNeRFFL, zhou2023feature,Chen2023GaussianEditorSA} typically rely on distilling features from 2D foundation models and assign additional attributes to each voxel or Gaussian for task-specific purposes. In contrast, our work explores how to manipulate scenes based on a deeper, underlying structure beyond these low-level representations. While JacobiNeRF\citep{Xu_2023_CVPR} shares a similar concept of correlation shaping within the network, it focuses on label propagation and employs point-wise shaping, which limits its utility for diverse scene manipulations. Additionally, the mutual information calculation in ~\citet{Xu_2023_CVPR} is both memory- and time-intensive during optimization. Another approach involves using physics simulators to guide Gaussian editing ~\citep{xie2023physgaussian}, but it similarly encounters inefficiencies due to the high computational cost of solving partial differential equations at inference time. Our work builds on the idea of mutual information, offering more efficient optimization and enabling fast scene editing during inference by leveraging well-shaped correlations.

\paragraph{Scene representations for 3D editing} 
Despite advancements in reconstruction, 
interactive 3D scene editing 
remains challenging. 
Current neural representation-based approaches~\citep{wang2022clip,Yuan2022NeRFEditingGE,schwarz2020graf,Wang2023ProteusNeRFFL,zhou2023feature,Chen2023GaussianEditorSA} 
typically rely on attributes 
computed by distilling features 
from 2D foundation models 
and perform editing in a point-wise manner.
In contrast, 
our work delves into manipulating scenes based on a second-order correlation structure. 
The idea of enforcing second-order correlation is first proposed in 
JacobiNeRF~\citep{Xu_2023_CVPR}, 
while it mainly focuses on label propagation and one-time perturbation consistency, 
limiting its potential 
for diverse scene manipulations. 
Furthermore, 
the correlation shaping in~\citet{Xu_2023_CVPR} is both memory- and time-intensive during the training.
Other approaches involve 
using physics simulators to guide scene editing~\citep{Li2023PACNeRFPA,xie2023physgaussian}, 
yet incurring a high computational cost 
in solving partial differential equations at inference time. 
Our approach offers 
efficient optimization and enables 
fast scene editing during inference 
by learning continuously well-shaped tangent spaces.

\section{Modeling mutual information between 3D gaussians}\label{sec:method}

% \yang{
% a few questions that need to be answered to structure the writing of the method section:
% \begin{itemize}
%     \item why you choose gaussian splatting? especially those that encode the motion using an mlp
%     \yunchao{}
%     \item what is the drawback of such framework? and why this drawback needs us to pay attention?
%     \yunchao{Briefly speaking}
%     \item what are the current solutions for this drawback?
%     \item why the current solutions are not good enough?
%     \item what are the criteria that such pipeline (motion mlp) need to satisfy in order to resolves this issues?
%     \item your proposed solution, and why?
% \end{itemize}
% }

Our goal is to develop scene representations that not only capture 3D geometry and appearance with high precision but also encode the interrelationships between individual elements. These enriched representations enable efficient scene editing by ensuring coherent and realistic updates, where changes propagate in a coordinated and intuitive manner.

Building on 3D Gaussian Splatting (3DGS)~\citep{kerbl20233d}, a state-of-the-art technique for 3D reconstruction that represents scenes using independent sets of 3D Gaussians, we aim to introduce meaningful correlations between these Gaussians. To achieve this, we incorporate mutual information (MI) to capture their underlying interdependence. However, traditional MI-based methods involve heavy optimization processes and are typically applied point-wise, meaning they are valid only for the current representation. This renders them unsuitable for scene editing, which requires both real-time response and persistent editability.

To overcome this limitation, we propose a novel training approach that efficiently shapes the attribute decoding network to capture and maintain these correlations throughout the decoding process. Notably, these structured correlations persist even after multiple consecutive scene edits. Finally, we demonstrate how these well-formed correlations improve a range of downstream scene editing tasks.

\subsection{Preliminaries: 3D gaussian splatting and attibute decoding}
\label{sec:pre}

Given a dataset $\mathcal{D}$ consisting of multi-view 2D images and corresponding camera poses, 3D Gaussian Splatting (3DGS) reconstructs the 3D scene by learning a set of 3D Gaussians $\mathcal{G}=\{\mathbf{g}_1,\mathbf{g}_2,...,\mathbf{g}_N\}$, where $N$ denotes the number of 3D Gaussians in the entire scene. Each 3D Gaussian $\mathbf{g}_i$ possesses multiple attributes, denoted as $\{\mathbf{x}_i, \mathbf{s}_i, \mathbf{q}_i, \mathbf{o}_i, \mathbf{c}_i\}$. Specifically, $\mathbf{x}_i \in \mathbb{R}^3$ represents the location of the centroid of $\mathbf{g}_i$, $\mathbf{s}_i \in \mathbb{R}^3$ denotes the scale, and $\mathbf{q}_i \in \mathbb{R}^4$ denotes the rotational quaternion, such that $\mathbf{s}_i$ and $\mathbf{q}_i$ jointly represent the 3D covariance matrix $\mathbf{\Sigma}$ of $\mathbf{g}_i$. Additionally, $\mathbf{o}_i \in \mathbb{R}$ represents the opacity, and $\mathbf{c}_i$ denotes the color, which is represented as the spherical harmonics (SH) coefficients. Given a camera pose, the color $C(p)$ of a pixel $p$ can be computed by blending a set of depth-ordered Gaussians $\mathcal{N}$ that overlap with $p$: 
\begin{align} 
C(p) = \sum_{i\in\mathcal{N}}{\mathbf{c}i\alpha_i\prod_{j=1}^{i-1}(1-\alpha_j)}, 
\end{align} where $\mathbf{c}_i$ is the color of the $i$-th Gaussian, and $\alpha_i$ is the blending weight, calculated using the opacity, projected 2D covariance of the Gaussian, and the location of the pixel.

While 3DGS offers several advantages, its sparse and unorganized structure leads to significant storage demands and makes scalability challenging. To address this, mutual relationships between Gaussians have been explored in subsequent works to eliminate structural redundancies. ~\citet{Lu2023ScaffoldGSS3,lee2024compact} proposed leveraging spatial relationships (i.e., features of Gaussians in a local region tend to have similar values) by employing neural networks to decode their features: 
\begin{align} 
a(\mathbf{g}_i) = \mathbf{\Phi}_a(f(\mathbf{g}_i), d;\theta), \end{align} 
where $a(\mathbf{g}_i)$ denotes attributes such as the color $\mathbf{c}_i$ of the Gaussian $\mathbf{g}_i$. The Gaussian attributes are decoded by an attribute decoding network $\mathbf{\Phi}_a$ with parameters $\theta$, taking the feature of the Gaussian $f(\mathbf{g}_i)$ and the viewing direction $d$ as inputs. Incorporating the viewing direction into inputs improves the scene representation’s robustness to substantial viewpoint changes and lighting effects.

Although associating millions of Gaussians via a network has proven effective, we argue that deeper exploration of the correlation structure of Gaussians through mutual information can further enhance scene editing, enabling better coordination among Gaussians for more effective results.

\subsection{Mutual information between 3D gaussians}
\label{sec:MI}

Traditional 3DGS losses predominantly focus on first-order supervision, optimizing for view synthesis rather than scene interaction. In contrast, this work introduces second-order supervision to shape the attribute decoding multilayer perceptron (MLP), facilitating efficient scene editing. Specifically, we optimize the network parameters to induce a global structure that encodes correlations among individual Gaussians. This ensures that modifications to the attribute decoding network's parameters, denoted by $\theta$, directly correspond to meaningful scene-editing operations. As a result, network outputs produce coherent and semantically consistent editing effects in the rendered scene, driven by the learned co-variation correlations.

To formalize this, we study the correlations between Gaussians under representation parameter changes in the output attributes. We analyze this behavior by considering two randomly selected Gaussians, $\mathbf{g}_i$ and $\mathbf{g}_j$, and their corresponding offsets in output attributes, $\hat{a}(\mathbf{g}_i)$ and $\hat{a}(\mathbf{g}_j)$. These attribute changes arise from the perturbed network $\hat{F}_a(\cdot; \theta^D + \mathbf{n})$, where $\mathbf{n} \in \mathbb{R}^D$ represents a perturbation applied to the parameter set $\theta^D$. 

Following ~\citet{Xu_2023_CVPR}, we show that the mutual information between the attribute variations $\hat{a}(\mathbf{g}_i)$ and $\hat{a}(\mathbf{g}_j)$ is proportional to the cosine similarity of the Jacobians of ${\mathbf{\Phi}_a}$ with respect to the perturbed parameters $\theta_D$: \begin{equation} 
\label{eq:gradient} 
\mathbb{I}(\hat{a}(\mathbf{g}_i),\hat{a}(\mathbf{g}_j)) = \log\left(\frac{1}{\sqrt{1-\cos^2(\gamma)}}\right) + \text{const.}, 
\end{equation} 
where $\gamma$ denotes the angle between the Jacobians $\frac{\partial{\mathbf{\Phi}_a(f(\mathbf{g}_i,d);\theta)}}{\partial{\theta_D}}$ and $\frac{\partial{\mathbf{\Phi}_a(f(\mathbf{g}_j),d;\theta)}}{\partial{\theta_D}}$. 
For clarity, we denote the Jacobians as $\partial{\mathbf{\Phi}}_i$ and $\partial{\mathbf{\Phi}}_j$. This formulation reveals that the mutual information $\mathbb{I}(\hat{a}(\mathbf{g}_i), \hat{a}(\mathbf{g}_j))$ is positively correlated with the cosine similarity of their gradients with respect to the perturbed parameters.

\begin{wrapfigure}[14]{r}{.5\textwidth}
\vspace{-4ex}
\centering
\begin{center}
 \includegraphics[width=\linewidth]{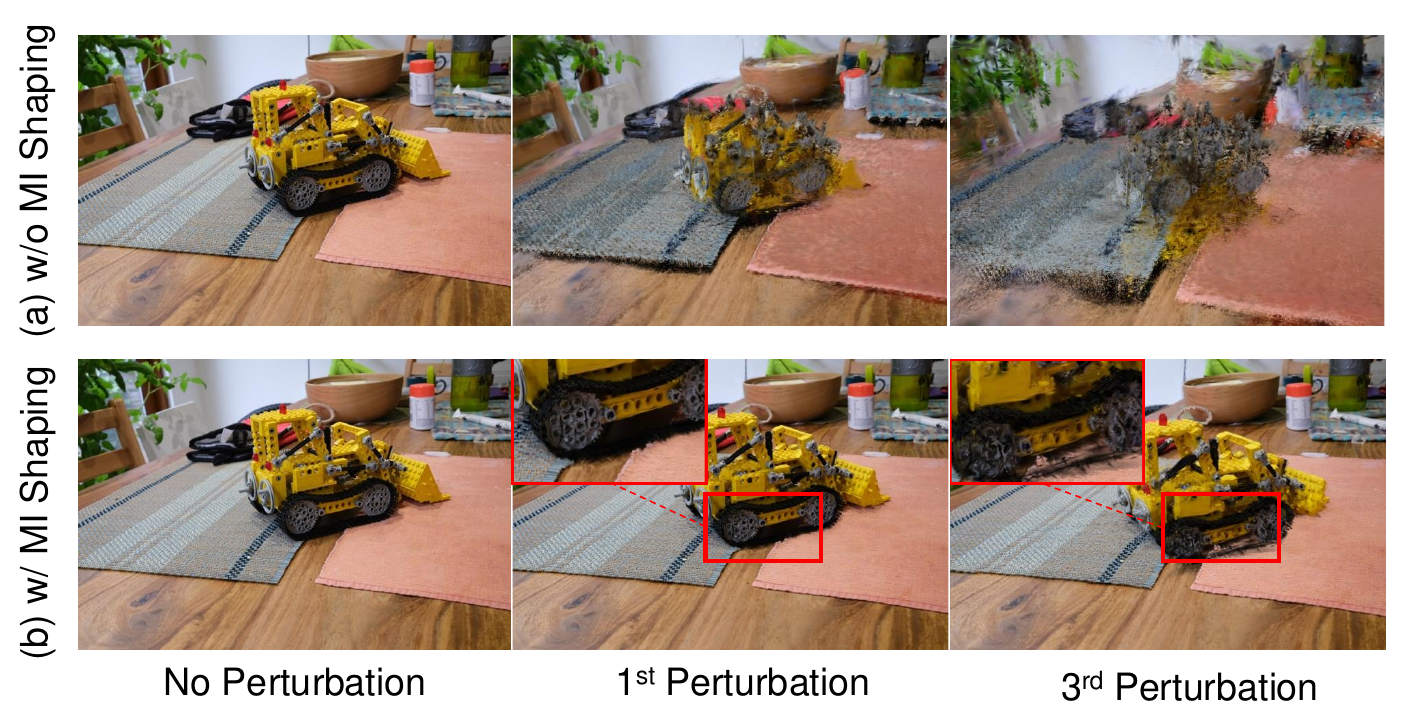}
 \vspace{-5ex}
  \caption{Perturb the attribute decoding network by the Jacobian of a Gaussian in the bulldozer without (a) or with (b) MI shaping~\citep{Xu_2023_CVPR}, and then move Gaussians according to similarities of Jacobian with selected one.}
\label{fig:prob}
\end{center}
\end{wrapfigure}

We apply Eq.~\ref{eq:gradient} to evaluate whether the correlations between Gaussians are well-structured by the attribute decoding network $\mathbf{\Phi}_a$. As shown in Fig.~\ref{fig:prob} (a), network weights optimized solely by the reconstruction loss fail to reveal meaningful correlations between Gaussians, primarily due to the unstructured nature of 3DGS. Despite the association of millions of Gaussians via the decoding network, they remain disorganized— a limitation of methods optimized solely for novel view synthesis.

JacobiNeRF~\citep{Xu_2023_CVPR} addresses this by enforcing perturbation consistency through maximizing the quantity in Eq.~\ref{eq:gradient}, enabling label propagation tasks. However, this approach is not suitable for scene editing in the 3DGS framework for two main reasons: 
(1) It does not support consecutive scene editing. As shown in Fig.~\ref{fig:prob} (b), previous MI shaping fails after multiple editing operations, leading to unrealistic scene updates. This is because the shaping in \citet{Xu_2023_CVPR} is \textit{point-wise}: while $\partial{\mathbf{\Phi}}_i$ and $\partial{\mathbf{\Phi}}_j$ may be similar in the original tangent space, there is no guarantee that they remain similar after parameter perturbation. (2) The optimization process is complex, requiring substantial storage to maintain the computational graph and significant time for Jacobian computation. This approach is inefficient for 3DGS and fails to meet the real-time demands of user-driven scene interactions. 

Next, we will discuss how to ensure consistency across consecutive parameter changes while maintaining training efficiency.

\subsection{Mutual information shaping with consistency and efficiency}
\label{sec:activation}

To ensure consistency in correlation shaping after perturbation, we begin by formulating the gradient expression for an in-depth analysis. 
Given that attribute decoding networks in neural fields (e.g., NeRFs and 3DGS) are MLPs, we focus on perturbing the weights $W^{(l)}$ of the $l$-th linear layer in the network $\mathbf{\Phi}_a$ (i.e., $\theta_D = W^{(l)}$), expressed as $h^{(l)}=W^{(l)}\sigma(h^{(l-1)})+b^{(l)}$. 
Here, $\sigma$ denotes the non-linear activation function, $b^{(l)}$ is the bias, and $h^{(l)}$ is the (hidden) output of the $l$-th layer.

To reveal the relationship between the gradient of a Gaussian $\mathbf{g}_i$, $\partial \mathbf{\Phi}_i = \frac{\partial \mathbf{\Phi}(f(\mathbf{g}_i, d);\theta)}{\partial \theta_D}$, and the perturbed weight $W^{(l)}$, we consider the gradient of an arbitrary scalar output $z$ in the attribute decoding network with respect to $W^{(l)}$: \begin{align} z = Y(h^{(l)})=Y(W^{(l)}\sigma(h^{(l-1)})+b^{(l)}), \end{align} where $Y$ denotes all transformations after the $l$-th layer, and $z$ is an arbitrary scalar in the vector output of $\mathbf{\Phi}$. The differential $\mathrm{d}{z}$ can be written as: \begin{equation} \begin{aligned} \mathrm{d}{z}&=\frac{\partial z}{\partial h^{(l)}}^\top\mathrm{d}{h^{(l)}}=\sigma(h^{(l-1)})\frac{\partial z}{\partial h^{(l)}}^\top\mathrm{d}{W^{(l)}} =\text{tr}(\sigma(h^{(l-1)})\frac{\partial z}{\partial h^{(l)}}^\top\mathrm{d}{W^{(l)}}), \end{aligned} \end{equation} since $\mathrm{d}{z}=\text{tr}\left(\frac{\partial z}{\partial W^{(l)}}^\top\mathrm{d}{W^{(l)}}\right)$, we can derive: \begin{align} 
\label{eq:trans} 
\frac{\partial z}{\partial W^{(l)}}=\frac{\partial z}{\partial h^{(l)}}\sigma(h^{(l-1)})^\top. 
\end{align} 
By expressing $\cos(\gamma)$ in Eq.~\ref{eq:gradient}
 using the right-hand side of Eq.~\ref{eq:trans}, we derive: 
\begin{equation} 
\label{eq:activation}
\cos(\partial{\mathbf{\Phi}^{(d)}_i},\partial{\mathbf{\Phi}^{(d)}_j}) \approx \cos (\partial h^{(0)}_i,\partial h^{(0)}_j), \quad d \in \mathbb{N}, \end{equation} when $\partial h^{(0)}_i$ and $\partial h^{(0)}_j$ point in the same or opposite directions. Here, $\partial{\mathbf{\Phi}^{(d)}_i}$ represents the Jacobian after the $d$-th perturbation, and $\partial h^{(0)}_i$ represents the Jacobian $\frac{\partial h^{(l)}}{\partial W^{(l)}}$ of $\mathbf{g}_i$ without perturbation. Upon further derivation, we find that $\partial h$ corresponds to repeated activations $\sigma(h^{(l-1)})$. {A detailed derivation and the proof of Eq.~\ref{eq:activation} can be found in the Appendix C}. Eq.~\ref{eq:activation} indicates that activation shaping can substitute for Jacobian calculation in Eq.~\ref{eq:activation} to satisfy object-level mutual relations, i.e., $\lVert\cos(\gamma)\rVert$ approaches 1 if $\mathbf{g}_i$ and $\mathbf{g}_j$ belong to the same object, or approaches 0 if not.

The key insight of Eq.~\ref{eq:activation} is that by optimizing activations instead of Jacobians, the correlations between Gaussians can be correctly shaped and remain consistent after successive parameter changes. This enables versatile editing operations in the scene and makes training much more efficient, as there is no need to maintain the computational graph for Jacobian calculation.

\subsection{Shaping correlations between gaussians for scene editing}
\label{sec:shape}
 \begin{figure}[!t]
    \includegraphics[width=\textwidth]{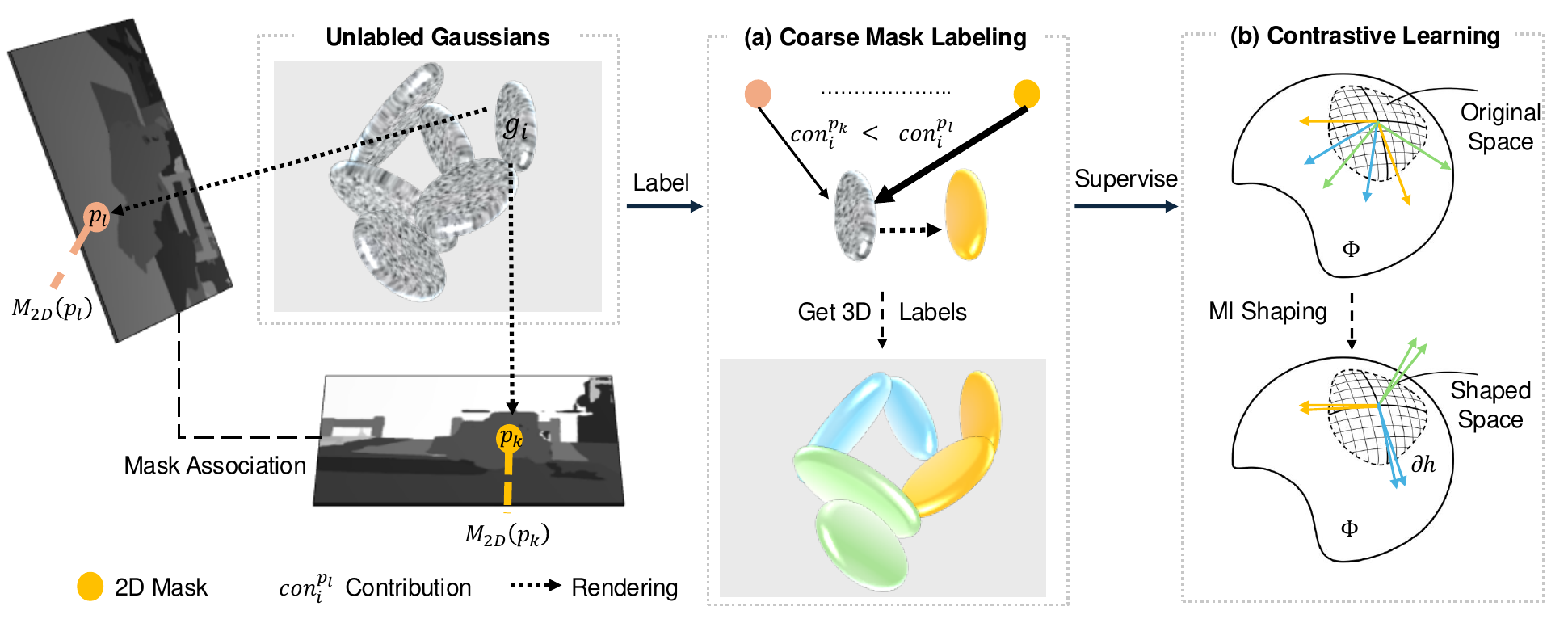}
    \caption{
    The training pipeline of our correlation shaping: \textbf{(a)} We use SAM~\cite{kirillov2023segany} to generate 2D masks and a pre-trained zero-shot tracker~\cite{cheng2023tracking} to associate masks from different views. We label the 3D mask of each Gaussian according to the 2D mask of the pixel that owns the Gaussian's maximal contribution during rendering across all views. \textbf{(b)} We use the labeled 3D masks as the supervision to conduct contrastive learning for mutual information shaping. After shaping, the Jacobians are consistently distributed in the tangent space.
    }
    \label{fig:method}
    \vspace{-2.5ex}
\end{figure}

According to Eq.~\ref{eq:gradient} and Eq.~\ref{eq:activation}, to ensure that two 3D Gaussians remain highly correlated consistently even after parameter changes -- thus supporting a sequence of editing operations without reshaping the parameters -- we shape their activations $\partial h$ within the attribute decoding network $\mathbf{\Phi}a$. To enable meaningful scene editing, we define positive and negative pairs based on object-level relationships, where $(g_i, g_i^+)$ represents a pair of Gaussians belonging to the same entity, and $(g_i, g_i^-)$ are independent Gaussians. For positive pairs, we align their activations in $\mathbf{\Phi}a$ relative to the perturbed layer $W^{(l)}$, while for negative pairs with minimal mutual information, we enforce their $\partial h$ to be orthogonal. To achieve this, we minimize the InfoNCE loss~\citep{Oord2018RepresentationLW}: 
\begin{equation} \label{eq:infonce} 
\mathcal{L}_{\text{MI}} = - \log \frac{\exp(| \cos(\partial h_i, \partial h_{i^+}) | / \tau)}{\sum_{i^+ \cup \{i^-\}} \exp(| \cos(\partial h_i, \partial h_{i^-}) | / \tau)}, 
\end{equation} 
where $\tau$ is the temperature parameter. This loss encourages highly correlated Gaussians to exhibit large activation similarity while ensuring that uncorrelated Gaussians exhibit near-zero similarity.

\subsection{{Supervision refinement}}

We now address how to sample positive and negative pairs from all 3D Gaussians. Identifying the label of each 3D Gaussian is challenging, as 3DGS is trained from a set of 2D posed images, complicating the recovery of the joint 3D distribution. However, by leveraging the advances of 2D vision foundation models~\citep{kirillov2023segany}, we label a coarse mask by lifting 2D mask to 3D: 
\begin{equation}
\label{eq:mask} 
M_{3D}(g_i)=M_{2D} (\mathop{\arg\max}\limits_{p} \, con^p_i),
\end{equation} 
where $con^p_i$ represents the contribution of $g_i$ to pixel $p$ during rendering, and $M_{3D}$ and $M_{2D}$ denote the 3D mask of Gaussians and the 2D mask of pixels. Eq.~\ref{eq:mask} indicates that the mask of a 3D Gaussian $g_i$ is determined by the 2D mask of a pixel where $g_i$ contributes the most across all 2D images. We compute the contribution $con^p_i$ based on a trained 3DGS that contains geometric priors of the scene. Thus the mask of $g_i$ is likely to be determined by the 2D mask where its maximal contribution appears. In order to harmonize 2D mask IDs produced from different views, we use a pre-trained zero-shot tracker~\citep{cheng2023tracking} to associate 2D masks by treating them as a video sequence.
We also provide a version without using tracker in the experiments, and it is elaborated in the Appendix.

To further enhance efficiency, 
we introduce a smoothness regularization 
in the self-supervised training. 
Since objects are continuous in space, neighboring Gaussians of $\mathbf{g}i$ are likely to belong to the same object as $\mathbf{g}_i$. This observation helps shape the Jacobians of 3D Gaussians that are heavily occluded by others and may not be accurately annotated in the coarse 3D segmentation. Such Gaussians can be supervised by nearby, well-labeled Gaussians. Specifically, $\mathcal{L}_{R}$ is formulated as: 
\begin{equation} \label{eq:reg} 
\mathcal{L}_{R}=\mathbb{E}_{\mathbb{P}}[\frac{1}{k}\sum_{x^+ \in n(x)}1.0-\cos(\mathbf{\Phi}_x, \mathbf{\Phi}_{x^+})],
\end{equation} 
where $\mathbb{P}$ denotes the input space, $n(x)$ represents the neighboring Gaussians of $x$, and $k=\lVert n(x) \rVert$ is the number of neighbors. In practice, the distance between two Gaussians is defined by the Euclidean distance between their centroids $\mathbf{x}$, and $k$ is a fixed hyperparameter during training.

\subsection{Full shaping pipeline}
Fig.~\ref{fig:method} illustrates our training pipeline, and the total loss as follows: 
\begin{equation} 
\label{eq:loss} 
\mathcal{L}=\mathcal{L}_{3DGS}+\lambda_{m}\mathcal{L}_{MI}+\lambda_{r}\mathcal{L}_{R},
\end{equation} 
where $\mathcal{L}_{3DGS}$ is the photometric reconstruction loss from the original 3DGS pipeline, $\mathcal{L}_{MI}$ is the contrastive learning loss (Eq.~\ref{eq:infonce}), and $\mathcal{L}_{R}$ is the regularization loss, with $\lambda_{r}$ representing its corresponding weight. 
% In practice, we first train a baseline 3DGS for reconstruction, then apply the mixed loss in Eq.~\ref{eq:loss} during the fine-tuning stage.

\subsection{Correlation-based scene editing}
\label{sec:edit}
After shaping the correlations, scene editing is performed by perturbing the parameters of the attribute decoding network. Specifically, we perturb the attribute decoding network by applying the average Jacobian $\partial \mathbf{\Phi}_a$ of user-selected Gaussians. When editing through a 2D image rather than directly in 3D space, we select the Gaussians that have the largest influence on the color of the user-selected pixels, similar to the contribution calculation in Eq.~\ref{eq:mask}.

Different branches of the network produce distinct editing effects. For object removal, we perturb the opacity branch $\mathbf{\Phi}_o$ in the gradient direction that reduces output opacity. This can drive the opacity of the relevant Gaussians to zero, effectively removing the object in the rendered scene. For object re-colorization, we perturb the color branch $\mathbf{\Phi}_c$ along different gradient directions of output channels to generate diverse color changes. 
In the case of 3D segmentation and object movement, we move Gaussians according to similarities in Jacobian with the selected one. When extending to dynamic scenes~\citep{Pumarola2020DNeRFNR}, we achieve object movement by shaping and perturbing the motion network. More editing tasks including 3D object style transfer are discussed in the Appendix.
\section{Experiments}
\subsection{Implementation details}
\textbf{Training.} We shape mutual information between Gaussians starting from a pre-trained 3D Gaussian Splatting, applying a fine-tuning stage for optimizing attribute decoding network $\mathbf{\Phi}_a$. We follow the shaping pipeline in Fig.~\ref{fig:method} for $1500$ iterations. The finetuning stage is conducted on a single RTX 3090 GPU for about 1 minute. Detailed hyper-parameter settings can be found in the Appendix.

\textbf{Datasets.} We test our method on open-world datasets LERF-Localization~\citep{lerf2023}, derived LERF-Mask dataset~\citep{ye2023gaussian} with ground-truth segmentation, and Mip-NeRF 360~\citep{barron2022mip}, in order to evaluate the segmentation and editing quality in complex and compositional scenarios. We also test the effectiveness of our shaping on the real dynamic scenes in the D-NeRF dataset~\citep{pumarola2021d}. Additionally, we provide results on the outdoor unbounded scenes in dataset NERDS 360~\citep{irshad2023neo} in the Appendix.

% \textbf{Baselines.} 
% To evaluate the efficiency and accuracy of our correlation shaping for scene editing, we utilize the MI shaping loss in JacobiNeRF~\citep{Xu_2023_CVPR} as a substitute for our MI shaping $\mathcal{L}_{MI}$ in Eq.~\ref{eq:loss}, which is a point-wise method for shaping the network in response to one-time perturbation. While keeping other loss terms the same for a fair comparison, we name the baseline \textit{JaocbiGS}. Besides JacobiGS, we adopt three more baselines: LERF~\cite{lerf2023}, Gaussian Grouping~\citep{ye2023gaussian} and SAGA~\citep{cen2023segment} that do not use MI shaping to achieve segmentation. These methods assign an extra feature for each volume or 3D Gaussian to predict the 2D mask through projection.

\subsection{3D segmentation}
\label{sec:seg}
After shaping the correlations 
among Gaussians, 
our approach enables 3D segmentation by simply perturbing the network parameters.
We use Grounding DINO~\citep{liu2023grounding} to generate the 2D mask prompt. We compare our method with the state-of-the-art open vocabulary 3D segmentation methods, including both the NeRF-based~\citep{lerf2023,cen2023segment}, and 3DGS-based methods~\citep{Xu_2023_CVPR, ye2023gaussian}. {It's worth noting that Gaussian Grouping~\citep{ye2023gaussian} use the consistent 2D masks generated from the video tracker}. To evaluate the effectiveness of our improvement on correlation shaping, we utilize the MI shaping loss in JacobiNeRF~\citep{Xu_2023_CVPR} as a substitute for our MI shaping $\mathcal{L}_{MI}$ in Eq.~\ref{eq:loss}, which only permits one-time perturbation consistency.
While keeping other loss terms {and supervision refinement} the same for a fair comparison, we name the baseline \textit{JaocbiGS}. We also include the results of using foundation models.

\begin{figure}[!t]
    \includegraphics[width=\textwidth]{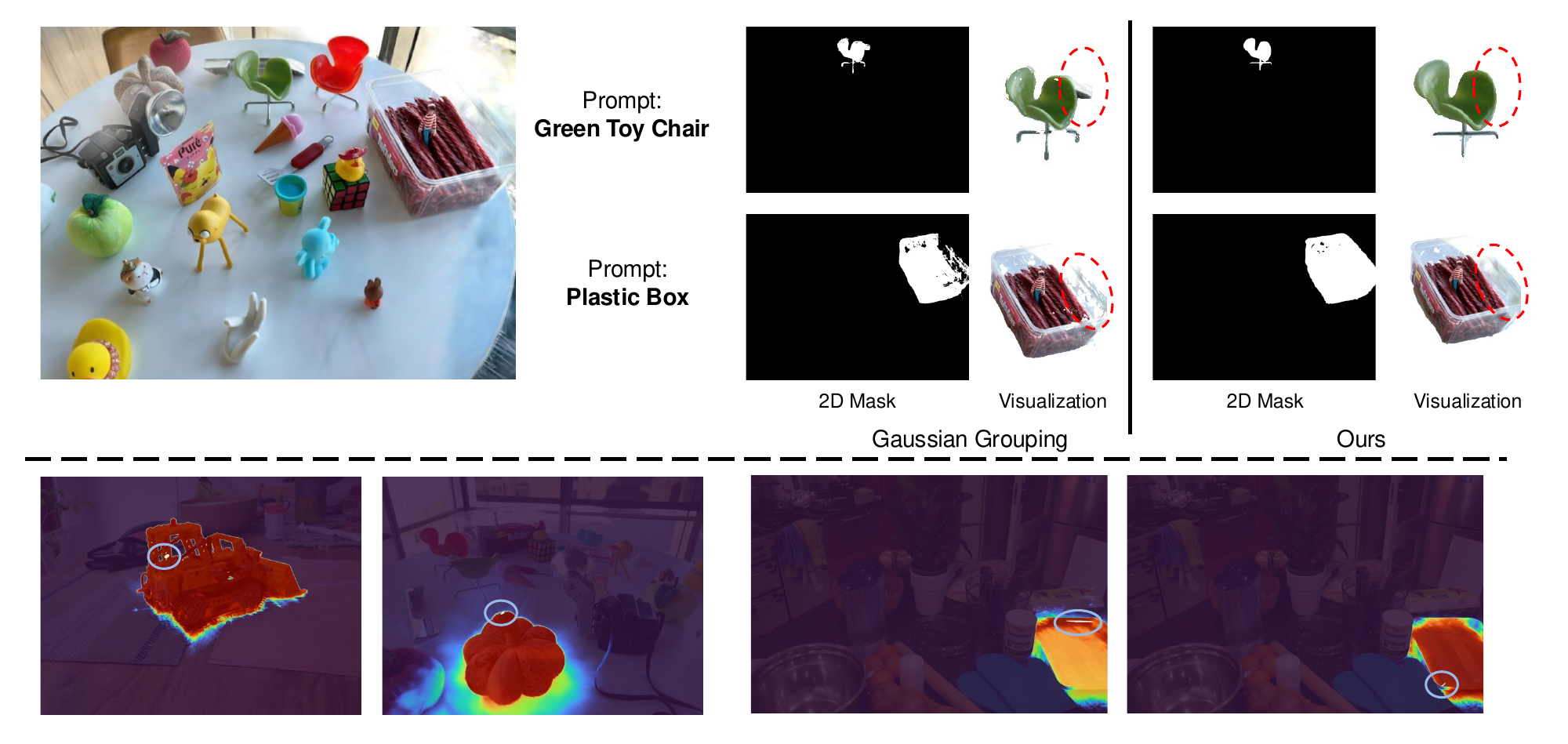}
    \vspace{-18pt}
    \caption{Top: 
    qualitative results of open vocabulary segmentation~\cite{ye2023gaussian}. Bottom: gallery of relevance maps when perturbing a single Gaussian (highlighted in the purple circle).}
    \label{fig:seg}
    \vspace{-10pt}
\end{figure}

For quantitative results, Tab.~\ref{table:seg} demonstrates that ours significantly outperforms than both NeRF-based and GS-based methods. Compared with previous state-of-the-art results, our method increases the mIoU by averagely $11\%$. Qualitative results are shown in Fig.~\ref{fig:seg}. 
We can see that our method produces sharp boundaries for the segmentation without less blurs. 
We also provide the relevance map when perturbing a single Gaussian to select the object. The relevance scores are represented as the cosine similarities of the Jacobians $\partial\mathbf{\Phi}$ of the perturbed Gaussian and all other Gaussians. As in Fig.~\ref{fig:seg}, when perturbing a single Gaussian, 
the corresponding object is highlighted, evidencing ours capability of segmentation from another viewpoint. 
It also shows that the relevance of Jacobian would be higher if a Gaussian is closer to the perturbed one. These phenomena comply with our expectations, indicating the effectiveness of the proposed MI shaping.

\begin{table}[!t]
\centering
\caption{Quantitative results of Open Vocabulary Segmentation on LERF-Mask dataset. The time and memory costs mentioned refer to the total consumption of training and finetuning (if have). Comparisons with non-3DGS method in this table are not feasible.}
\resizebox{\textwidth}{!}{
\begin{tabular}{ccccccc|ccc} \toprule
\multirow{2}*{Model} & \multicolumn{2}{c}{figurines} & \multicolumn{2}{c}{ramen} & \multicolumn{2}{c}{teatime} & \multicolumn{3}{c}{Average Training Cost}\\
~ & mIoU & mBIoU& mIoU & mBIoU & mIoU & mBIoU & Time/Minutes & Memory/M & \#GS Used\\ \midrule
{JacobiNeRF~\citep{Xu_2023_CVPR}}& {22.4} & {27.6} & {7.2} & {6.9} & {42.6} & {36.9} & {*} & {*} & {*}\\
LERF~\citep{lerf2023}& 33.5 & 30.6 & 28.3 & 14.7 & 49.7 & 42.6 & * & * & *\\
DEVA~\citep{cheng2023tracking} & 46.2 & 45.1 & 56.8 & 51.1 & 54.3 & 52.2 & * & * & * \\
SA3D~\citep{cen2023segment}& 24.9 & 23.8 & 7.4 & 7.0 & 42.5 & 39.2 & * & * & * \\
JacobiGS~\cite{Xu_2023_CVPR}& 62.2 & 60.1 & 64.3 & 57.7 & 69.7 & 70.1 & 30.2 & 160.2 & 17\%\\
Gaussian Grouping~\cite{ye2023gaussian}& 69.7 & 67.9 & 77.0 & 68.7 & 71.7 & 66.1 & 55.2 & 757.2 & 100\%\\
\cdashline{1-10}
{Ours w/o DEVA \& Regularization}& {72.2} & {69.8} & {79.5} & {69.9} & {77.2} & {70.1} & {19.7} & {160.2} & {15\%}\\
Ours w/o DEVA & 72.5 & 70.3 & 79.1 & 69.9 & 77.8 & 70.2 & 19.7 & 160.2 & 15\% \\
Ours & \textbf{80.6}& \textbf{77.2}& \textbf{80.6}& \textbf{70.1}& \textbf{84.5}& \textbf{78.7} & \textbf{16.3}& \textbf{160.2} & \textbf{7\%}\\
\bottomrule
\end{tabular}
}
% \vspace{0.5ex}
\label{table:seg}
\end{table}

The superior performance of our methods can be attributed to the learned awareness of the scene structure.
Instead of directly assigning scene attributes to each Gaussian \cite{cen2023segment,zhou2023feature,ye2023gaussian}, we focus on shaping networks that act on the Gaussians in a way respecting the underlying structures. 
It helps a lot for the low-level representation like 3DGS when leveraged for the interactive tasks like segmentation and editting. 
The shaped MLP is aware of the entire scene structure rather than individual Gaussians. 
Notably, 
we only sample about $7\%$ of all the Gaussians during finetuning until convergence, 
boosting the training time while with little memory storage.

\subsection{Scene editing}
\label{exp:edit}

\begin{figure}[!t]
    \includegraphics[width=\textwidth]{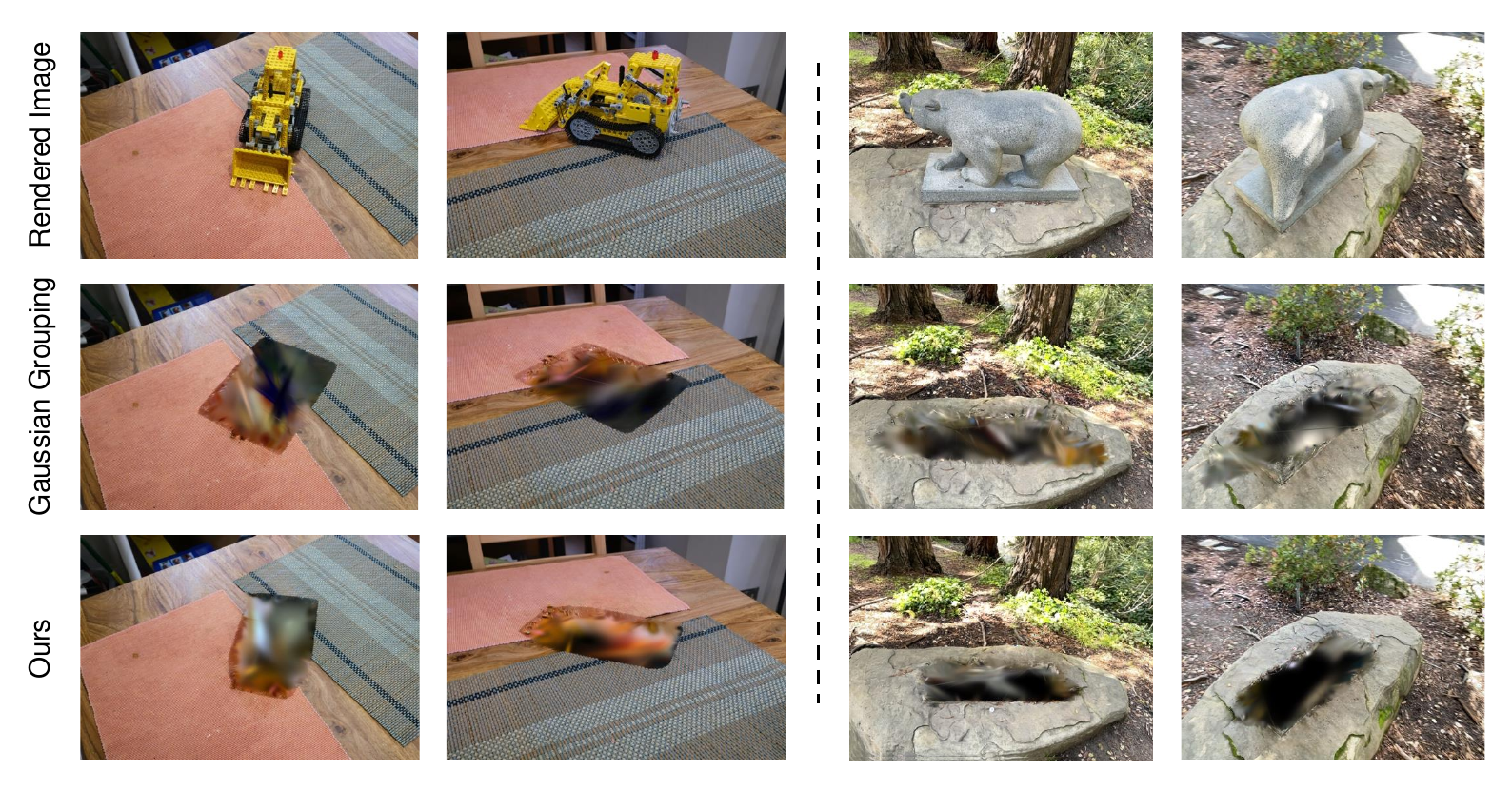}
    \vspace{-20pt}
    \caption{3D object removal on bear \& kitchen scenes. Compared to Gaussian Grouping, our method removes the object with a more fitting curve and less distortion of the irrelevant area.}
    \label{fig:remove}
    \vspace{-10pt}
\end{figure}

\begin{figure}[!t]
    \includegraphics[width=\textwidth]{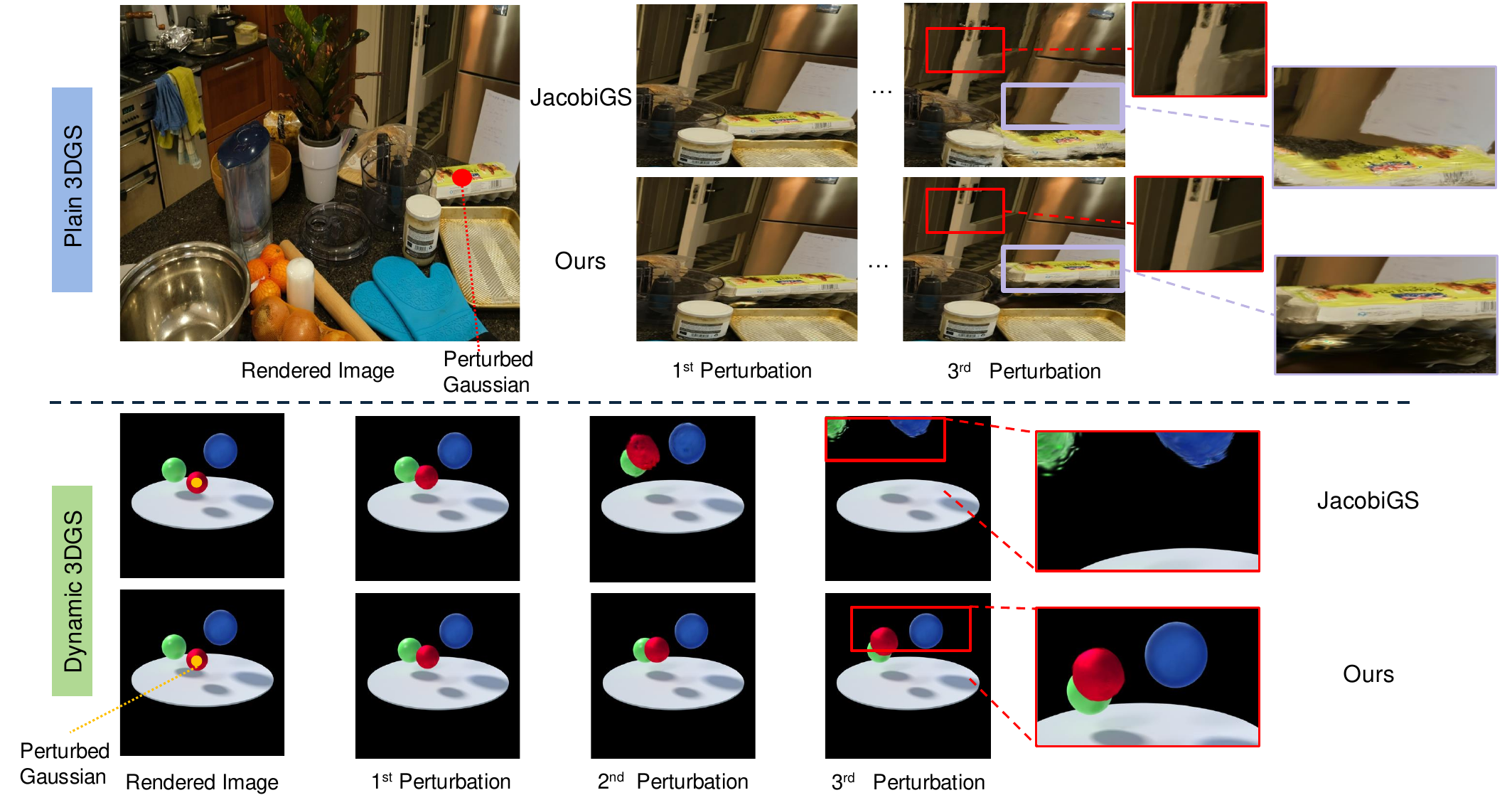}
    \vspace{-20pt}
    \caption{3D object movement on Mip-NeRF 360 ~\citep{barron2022mip} and D-NeRF ~\citep{Pumarola2020DNeRFNR}. Compared to the correlation shaping in ~\citet{Xu_2023_CVPR}, our method enables consecutive editing (no need for reshaping) on both static and dynamic scenes, while maintaining consistencies.}
    \label{fig:move}
    \vspace{-10pt}
\end{figure}

\textbf{3D Object removal.} 3D object removal is to delete the target object without affecting the background or unrelated area. The space behind the target is unknown from training views and thus it would be noisy after deletion. 
%Therefore, we expect the noisy area to be minimized. 
According to Fig.~\ref{fig:remove}, our method preserves more details of the unrelated area, and the contour of the area caused by the deletion fits the target object shape better. 
More explicitly, the noisy area resulted from our method after removal in the bear scene accurately matches the rectangle shape of the pedestal of the bear statue.  
While both are supervised by SAM and DEVA, our method separates the large object more clearly compared to Gaussian Grouping.

\textbf{3D Object movement.} We adopt the task of object movement to evaluate the editing quality under consecutive perturbations. {A single perturbation is to add the multiplication of the Jacobian $\partial \mathbf{\Phi}_i$ of selected Gaussian $g_i$ and a scaling factor $\sigma_s$ for controlling the impact}. As shown in Fig.~\ref{fig:move}, perturbing scenes with the same Gaussian reveals that our method induces reasonable motion (e.g., the carton is lifted up) while preserving the uncorrelated objects. It exhibits path-consistent motion after consecutive perturbations. In contrast, scenes generated by JacobiGS diverge into unrealistic representations after the second and third perturbation. For example, consecutive perturbations to the carton result in the distortion of the door behind the box (see the red boxes for highlights in Fig.~\ref{fig:move}). 
The observations demonstrate that our method supports continuous editing with consistency, thereby verifying the effectiveness of the proposed shaping technique.

\subsection{The effect on the reconstruction quality}

\begin{table}[!t]
\centering
\caption{Ablation study of correlation shaping.}
\resizebox{\textwidth}{!}{
\begin{tabular}{c|cccccc} \toprule
\multirow{2}*{Model} & \multicolumn{3}{c}{Mip-NeRF 360} & \multicolumn{3}{c}{D-NeRF}\\ \cmidrule(lr){2-4} \cmidrule(lr){5-7}
~  & PSNR$\uparrow$ & SSIM$\uparrow$ & LPIPS$\downarrow$ & PSNR$\uparrow$ & SSIM$\uparrow$ & LPIPS$\downarrow$ \\ \midrule
GS Baseline~\citep{kerbl20233d, yang2023deformable} & 28.69 & 0.870 & 0.182 & 41.0 & 0.995 & 0.009\\
Ours & 28.72 & 0.842 & 0.208 & 39.7 & 0.993 & 0.012\\
% Ours & 62.2 & 0.186 & 64.3 & 0.227 \\
\bottomrule
\end{tabular}
}
\label{table:ablation}
\vspace{0.9ex}
\end{table}

We conduct an ablation on how the correlation shaping affects the reconstruction quality. From Tab.~\ref{table:ablation}, we
can see that our framework has comparable performance to the baseline method on both static and dynamic scenes. There is a small PSNR drop on dynamics scenes, as we simply integrate our shaping loss into the original deformation network in \citet{yang2023deformable}, which aims for 4D reconstruction, 
making it a bit challenging to shape all timesteps,
and this is subject to future study.
% We evaluate our method on two perspectives: 1) If a Gaussian is perturbed, will all the Gaussians of the whole object move? We use \textit{acc} to calculate such accuracy. 2) If other Gaussians outside the object are also affected, the perturbation would create unrealistic scenarios. We further use \textit{LPIPS} to evaluate the produced image quality. Detailed evaluation could be found in the appendix.

% From Tab.~\ref{table:dynamics}, we could see that our method has better movement accuracy and image generation quality, indicating the naturalness of the generated motion. Furthermore, original JacobiNeRF shaping is totally supervised by 2D through projection, so here we ablate the coarse mask labeling from JacobiGS and follow the supervision in JaobiNeRF, we find that the performance deteriorates, further demonstrating the effectiveness of 3D supervision supported by coarse mask labeling.

% \begin{table}[!t]
% \centering
% \caption{.}
% \resizebox{\textwidth}{!}{
% \begin{tabular}{cccc} \toprule
% \multirow{2}*{Model} & \multicolumn{3}{c}{Bouncing ball} \\
% ~ & PSNR$\uparrow$ & SSIM\uparrow & LPIPS\downarrow \\ \midrule
% w/o InfoGS& 41.01 & 0.9953 & 0.0093\\
% w/ InfoGS& 37.62 & 0.9928 & 0.0153 \\
% \bottomrule
% \end{tabular}
% }
% % \vspace{0.5ex}
% \label{table:seg}
% \end{table}

% \subsection{Sparse Representations}

\section{Discussion}

% We investigate the challenge of shaping correlations between independent scene elements in a 3DGS model to enable scene editing through network parameter perturbations. Our derivation demonstrates that enforcing gradient alignment within the network ensures the consistency of mutual information between Gaussians, even after successive parameter perturbations. To achieve this, we introduce an efficient finetuning pipeline, incorporating a contrastive loss and a 2D-to-3D sampling strategy.

% With the well-shaped correlation structure, we illustrate how the resonance between Gaussians facilitates 3D segmentation and automatically directs millions of Gaussians to respond appropriately to diverse editing tasks. Despite this efficiency, our current pipeline has limitations when it comes to generating arbitrarily complex edits, particularly at a fine-grained level, such as object deformation. In future work, we aim to address this limitation by learning intrinsic object features and kinematics from large-scale video datasets, which would expand the system’s capabilities in more sophisticated scene manipulation tasks.

We attempt the challenge of shaping correlations between independent scene elements in a 3D Gaussian Scene (3DGS) model to enable scene editing through network parameter perturbations. Our derivation demonstrates that enforcing gradient alignment within the network ensures the consistency of mutual information between Gaussians, even after successive parameter perturbations. To achieve this, we introduce an efficient training pipeline, incorporating a contrastive loss and a 2D-to-3D sampling strategy.
With the well-shaped correlation structure, we illustrate how the resonance between Gaussians facilitates 3D segmentation and automatically directs millions of Gaussians to respond appropriately to accomplish diverse editing tasks. 
Despite this efficiency, our current pipeline has limitations when generating arbitrarily complex edits, particularly at a fine-grained level, such as object deformation. In future work, we aim to address this limitation by learning intrinsic object features and kinematics from large-scale video datasets, which would enhance the its capabilities in more sophisticated scene manipulation tasks.

% \section{Reproducibility statement}

% We have made significant efforts to ensure the reproducibility of our work. A code example is provided as supplementary material, demonstrating the core components of our approach. Upon acceptance, we will release all of the data and the complete training and testing code to facilitate the full reproducibility of our results.

\clearpage
\bibliographystyle{iclr2025_conference}
\bibliography{iclr2025_conference}

\begin{thebibliography}{52}
\providecommand{\natexlab}[1]{#1}
\providecommand{\url}[1]{\texttt{#1}}
\expandafter\ifx\csname urlstyle\endcsname\relax
  \providecommand{\doi}[1]{doi: #1}\else
  \providecommand{\doi}{doi: \begingroup \urlstyle{rm}\Url}\fi

\bibitem[Bao et~al.(2023)Bao, Zhang, Yang, Fan, Yang, Bao, Zhang, and Cui]{Bao2023SINESI}
Chong Bao, Yinda Zhang, Bangbang Yang, Tianxing Fan, Zesong Yang, Hujun Bao, Guofeng Zhang, and Zhaopeng Cui.
\newblock Sine: Semantic-driven image-based nerf editing with prior-guided editing field.
\newblock \emph{2023 IEEE/CVF Conference on Computer Vision and Pattern Recognition (CVPR)}, pp.\  20919--20929, 2023.
\newblock URL \url{https://api.semanticscholar.org/CorpusID:257687450}.

\bibitem[Barron et~al.(2022)Barron, Mildenhall, Verbin, Srinivasan, and Hedman]{barron2022mip}
Jonathan~T Barron, Ben Mildenhall, Dor Verbin, Pratul~P Srinivasan, and Peter Hedman.
\newblock Mip-nerf 360: Unbounded anti-aliased neural radiance fields.
\newblock In \emph{Proceedings of the IEEE/CVF Conference on Computer Vision and Pattern Recognition}, pp.\  5470--5479, 2022.

\bibitem[Bronson(1969)]{Bronson1969MatrixMA}
Richard Bronson.
\newblock Matrix methods: An introduction.
\newblock 1969.
\newblock URL \url{https://api.semanticscholar.org/CorpusID:117111125}.

\bibitem[Brooks et~al.(2023)Brooks, Holynski, and Efros]{brooks2023instructpix2pix}
Tim Brooks, Aleksander Holynski, and Alexei~A Efros.
\newblock Instructpix2pix: Learning to follow image editing instructions.
\newblock In \emph{Proceedings of the IEEE/CVF Conference on Computer Vision and Pattern Recognition}, pp.\  18392--18402, 2023.

\bibitem[Cen et~al.(2023)Cen, Zhou, Fang, Shen, Xie, Jiang, Zhang, Tian, et~al.]{cen2023segment}
Jiazhong Cen, Zanwei Zhou, Jiemin Fang, Wei Shen, Lingxi Xie, Dongsheng Jiang, Xiaopeng Zhang, Qi~Tian, et~al.
\newblock Segment anything in 3d with nerfs.
\newblock \emph{Advances in Neural Information Processing Systems}, 36:\penalty0 25971--25990, 2023.

\bibitem[Chen et~al.(2024)Chen, Wu, Cai, Harandi, and Lin]{Chen2024HACHA}
Yihang Chen, Qianyi Wu, Jianfei Cai, Mehrtash Harandi, and Weiyao Lin.
\newblock Hac: Hash-grid assisted context for 3d gaussian splatting compression.
\newblock \emph{ArXiv}, abs/2403.14530, 2024.
\newblock URL \url{https://api.semanticscholar.org/CorpusID:268553921}.

\bibitem[Chen et~al.(2023{\natexlab{a}})Chen, Chen, Zhang, Wang, Yang, Wang, Cai, Yang, Liu, and Lin]{Chen2023GaussianEditorSA}
Yiwen Chen, Zilong Chen, Chi Zhang, Feng Wang, Xiaofeng Yang, Yikai Wang, Zhongang Cai, Lei Yang, Huaping Liu, and Guosheng Lin.
\newblock Gaussianeditor: Swift and controllable 3d editing with gaussian splatting.
\newblock \emph{ArXiv}, abs/2311.14521, 2023{\natexlab{a}}.
\newblock URL \url{https://api.semanticscholar.org/CorpusID:265445359}.

\bibitem[Chen et~al.(2023{\natexlab{b}})Chen, Wang, and Liu]{Chen2023Textto3DUG}
Zilong Chen, Feng Wang, and Huaping Liu.
\newblock Text-to-3d using gaussian splatting.
\newblock \emph{ArXiv}, abs/2309.16585, 2023{\natexlab{b}}.
\newblock URL \url{https://api.semanticscholar.org/CorpusID:263139613}.

\bibitem[Cheng et~al.(2023)Cheng, Oh, Price, Schwing, and Lee]{cheng2023tracking}
Ho~Kei Cheng, Seoung~Wug Oh, Brian Price, Alexander Schwing, and Joon-Young Lee.
\newblock Tracking anything with decoupled video segmentation.
\newblock In \emph{ICCV}, 2023.

\bibitem[Deng et~al.(2022)Deng, Liu, Zhu, and Ramanan]{deng2022depth}
Kangle Deng, Andrew Liu, Jun-Yan Zhu, and Deva Ramanan.
\newblock Depth-supervised nerf: Fewer views and faster training for free.
\newblock In \emph{Proceedings of the IEEE/CVF Conference on Computer Vision and Pattern Recognition}, pp.\  12882--12891, 2022.

\bibitem[Fan et~al.(2023)Fan, Wang, Wen, Zhu, Xu, and Wang]{Fan2023LightGaussianU3}
Zhiwen Fan, Kevin Wang, Kairun Wen, Zehao Zhu, Dejia Xu, and Zhangyang Wang.
\newblock Lightgaussian: Unbounded 3d gaussian compression with 15x reduction and 200+ fps.
\newblock \emph{ArXiv}, abs/2311.17245, 2023.
\newblock URL \url{https://api.semanticscholar.org/CorpusID:265498655}.

\bibitem[Fridovich-Keil et~al.(2022)Fridovich-Keil, Yu, Tancik, Chen, Recht, and Kanazawa]{fridovich2022plenoxels}
Sara Fridovich-Keil, Alex Yu, Matthew Tancik, Qinhong Chen, Benjamin Recht, and Angjoo Kanazawa.
\newblock Plenoxels: Radiance fields without neural networks.
\newblock In \emph{Proceedings of the IEEE/CVF Conference on Computer Vision and Pattern Recognition}, pp.\  5501--5510, 2022.

\bibitem[Fu et~al.(2022)Fu, Zhang, Chen, Lu, Zhu, Zhou, Geiger, and Liao]{fu2022panoptic}
Xiao Fu, Shangzhan Zhang, Tianrun Chen, Yichong Lu, Lanyun Zhu, Xiaowei Zhou, Andreas Geiger, and Yiyi Liao.
\newblock Panoptic nerf: 3d-to-2d label transfer for panoptic urban scene segmentation.
\newblock In \emph{2022 International Conference on 3D Vision (3DV)}, pp.\  1--11. IEEE, 2022.

\bibitem[Gao et~al.(2022)Gao, Zhong, Xiang, Hong, Guo, and Zhang]{Gao2022ReconstructingPS}
Xuan Gao, Chenglai Zhong, Jun Xiang, Yang Hong, Yudong Guo, and Juyong Zhang.
\newblock Reconstructing personalized semantic facial nerf models from monocular video.
\newblock \emph{ACM Transactions on Graphics (TOG)}, 41:\penalty0 1 -- 12, 2022.
\newblock URL \url{https://api.semanticscholar.org/CorpusID:252846034}.

\bibitem[Garbin et~al.(2021)Garbin, Kowalski, Johnson, Shotton, and Valentin]{garbin2021fastnerf}
Stephan~J Garbin, Marek Kowalski, Matthew Johnson, Jamie Shotton, and Julien Valentin.
\newblock Fastnerf: High-fidelity neural rendering at 200fps.
\newblock In \emph{Proceedings of the IEEE/CVF International Conference on Computer Vision}, pp.\  14346--14355, 2021.

\bibitem[Irshad et~al.(2023)Irshad, Zakharov, Liu, Guizilini, Kollar, Gaidon, Kira, and Ambrus]{irshad2023neo}
Muhammad~Zubair Irshad, Sergey Zakharov, Katherine Liu, Vitor Guizilini, Thomas Kollar, Adrien Gaidon, Zsolt Kira, and Rares Ambrus.
\newblock Neo 360: Neural fields for sparse view synthesis of outdoor scenes.
\newblock In \emph{Proceedings of the IEEE/CVF International Conference on Computer Vision}, pp.\  9187--9198, 2023.

\bibitem[Karras et~al.(2021)Karras, Aittala, Laine, H{\"a}rk{\"o}nen, Hellsten, Lehtinen, and Aila]{karras2021alias}
Tero Karras, Miika Aittala, Samuli Laine, Erik H{\"a}rk{\"o}nen, Janne Hellsten, Jaakko Lehtinen, and Timo Aila.
\newblock Alias-free generative adversarial networks.
\newblock \emph{Advances in neural information processing systems}, 34:\penalty0 852--863, 2021.

\bibitem[Kerbl et~al.(2023)Kerbl, Kopanas, Leimk{\"u}hler, and Drettakis]{kerbl20233d}
Bernhard Kerbl, Georgios Kopanas, Thomas Leimk{\"u}hler, and George Drettakis.
\newblock 3d gaussian splatting for real-time radiance field rendering.
\newblock \emph{ACM Transactions on Graphics}, 42\penalty0 (4), 2023.

\bibitem[Kerr et~al.(2023)Kerr, Kim, Goldberg, Kanazawa, and Tancik]{lerf2023}
Justin Kerr, Chung~Min Kim, Ken Goldberg, Angjoo Kanazawa, and Matthew Tancik.
\newblock Lerf: Language embedded radiance fields.
\newblock In \emph{International Conference on Computer Vision (ICCV)}, 2023.

\bibitem[Kingma \& Ba(2014)Kingma and Ba]{kingma2014adam}
Diederik~P Kingma and Jimmy Ba.
\newblock Adam: A method for stochastic optimization.
\newblock \emph{arXiv preprint arXiv:1412.6980}, 2014.

\bibitem[Kirillov et~al.(2023)Kirillov, Mintun, Ravi, Mao, Rolland, Gustafson, Xiao, Whitehead, Berg, Lo, Doll{\'a}r, and Girshick]{kirillov2023segany}
Alexander Kirillov, Eric Mintun, Nikhila Ravi, Hanzi Mao, Chloe Rolland, Laura Gustafson, Tete Xiao, Spencer Whitehead, Alexander~C. Berg, Wan-Yen Lo, Piotr Doll{\'a}r, and Ross Girshick.
\newblock Segment anything.
\newblock \emph{arXiv:2304.02643}, 2023.

\bibitem[Lee et~al.(2024)Lee, Rho, Sun, Ko, and Park]{lee2024compact}
Joo~Chan Lee, Daniel Rho, Xiangyu Sun, Jong~Hwan Ko, and Eunbyung Park.
\newblock Compact 3d gaussian representation for radiance field.
\newblock In \emph{Proceedings of the IEEE/CVF Conference on Computer Vision and Pattern Recognition}, pp.\  21719--21728, 2024.

\bibitem[Li et~al.(2024)Li, Liu, and Zhou]{Li2024SGSSLAMSG}
Mingrui Li, Shuhong Liu, and Heng Zhou.
\newblock Sgs-slam: Semantic gaussian splatting for neural dense slam.
\newblock \emph{ArXiv}, abs/2402.03246, 2024.
\newblock URL \url{https://api.semanticscholar.org/CorpusID:267412978}.

\bibitem[Li et~al.(2023)Li, Qiao, Chen, Jatavallabhula, Lin, Jiang, and Gan]{Li2023PACNeRFPA}
Xuan Li, Yi-Ling Qiao, Peter~Yichen Chen, Krishna~Murthy Jatavallabhula, Ming Lin, Chenfanfu Jiang, and Chuang Gan.
\newblock Pac-nerf: Physics augmented continuum neural radiance fields for geometry-agnostic system identification.
\newblock \emph{ArXiv}, abs/2303.05512, 2023.
\newblock URL \url{https://api.semanticscholar.org/CorpusID:257427479}.

\bibitem[Liu et~al.(2020)Liu, Gu, Zaw~Lin, Chua, and Theobalt]{liu2020neural}
Lingjie Liu, Jiatao Gu, Kyaw Zaw~Lin, Tat-Seng Chua, and Christian Theobalt.
\newblock Neural sparse voxel fields.
\newblock \emph{Advances in Neural Information Processing Systems}, 33:\penalty0 15651--15663, 2020.

\bibitem[Liu et~al.(2023)Liu, Zeng, Ren, Li, Zhang, Yang, Li, Yang, Su, Zhu, et~al.]{liu2023grounding}
Shilong Liu, Zhaoyang Zeng, Tianhe Ren, Feng Li, Hao Zhang, Jie Yang, Chunyuan Li, Jianwei Yang, Hang Su, Jun Zhu, et~al.
\newblock Grounding dino: Marrying dino with grounded pre-training for open-set object detection.
\newblock \emph{arXiv preprint arXiv:2303.05499}, 2023.

\bibitem[Lu et~al.(2023)Lu, Yu, Xu, Xiangli, Wang, Lin, and Dai]{Lu2023ScaffoldGSS3}
Tao Lu, Mulin Yu, Linning Xu, Yuanbo Xiangli, Limin Wang, Dahua Lin, and Bo~Dai.
\newblock Scaffold-gs: Structured 3d gaussians for view-adaptive rendering.
\newblock \emph{ArXiv}, abs/2312.00109, 2023.
\newblock URL \url{https://api.semanticscholar.org/CorpusID:265551778}.

\bibitem[Martin-Brualla et~al.(2021)Martin-Brualla, Radwan, Sajjadi, Barron, Dosovitskiy, and Duckworth]{martin2021nerf}
Ricardo Martin-Brualla, Noha Radwan, Mehdi~SM Sajjadi, Jonathan~T Barron, Alexey Dosovitskiy, and Daniel Duckworth.
\newblock Nerf in the wild: Neural radiance fields for unconstrained photo collections.
\newblock In \emph{Proceedings of the IEEE/CVF Conference on Computer Vision and Pattern Recognition}, pp.\  7210--7219, 2021.

\bibitem[Mildenhall et~al.(2021)Mildenhall, Srinivasan, Tancik, Barron, Ramamoorthi, and Ng]{mildenhall2021nerf}
Ben Mildenhall, Pratul~P Srinivasan, Matthew Tancik, Jonathan~T Barron, Ravi Ramamoorthi, and Ren Ng.
\newblock Nerf: Representing scenes as neural radiance fields for view synthesis.
\newblock \emph{Communications of the ACM}, 65\penalty0 (1):\penalty0 99--106, 2021.

\bibitem[M{\"u}ller et~al.(2022)M{\"u}ller, Evans, Schied, and Keller]{muller2022instant}
Thomas M{\"u}ller, Alex Evans, Christoph Schied, and Alexander Keller.
\newblock Instant neural graphics primitives with a multiresolution hash encoding.
\newblock \emph{ACM Transactions on Graphics (ToG)}, 41\penalty0 (4):\penalty0 1--15, 2022.

\bibitem[{\"O}zye{\c{s}}il et~al.(2017){\"O}zye{\c{s}}il, Voroninski, Basri, and Singer]{ozyecsil2017survey}
Onur {\"O}zye{\c{s}}il, Vladislav Voroninski, Ronen Basri, and Amit Singer.
\newblock A survey of structure from motion*.
\newblock \emph{Acta Numerica}, 26:\penalty0 305--364, 2017.

\bibitem[Pumarola et~al.(2020)Pumarola, Corona, Pons-Moll, and Moreno-Noguer]{Pumarola2020DNeRFNR}
Albert Pumarola, Enric Corona, Gerard Pons-Moll, and Francesc Moreno-Noguer.
\newblock D-nerf: Neural radiance fields for dynamic scenes.
\newblock \emph{2021 IEEE/CVF Conference on Computer Vision and Pattern Recognition (CVPR)}, pp.\  10313--10322, 2020.
\newblock URL \url{https://api.semanticscholar.org/CorpusID:227227965}.

\bibitem[Pumarola et~al.(2021)Pumarola, Corona, Pons-Moll, and Moreno-Noguer]{pumarola2021d}
Albert Pumarola, Enric Corona, Gerard Pons-Moll, and Francesc Moreno-Noguer.
\newblock D-nerf: Neural radiance fields for dynamic scenes.
\newblock In \emph{Proceedings of the IEEE/CVF Conference on Computer Vision and Pattern Recognition}, pp.\  10318--10327, 2021.

\bibitem[Reiser et~al.(2021)Reiser, Peng, Liao, and Geiger]{Reiser2021KiloNeRFSU}
Christian Reiser, Songyou Peng, Yiyi Liao, and Andreas Geiger.
\newblock Kilonerf: Speeding up neural radiance fields with thousands of tiny mlps.
\newblock \emph{2021 IEEE/CVF International Conference on Computer Vision (ICCV)}, pp.\  14315--14325, 2021.
\newblock URL \url{https://api.semanticscholar.org/CorpusID:232352619}.

\bibitem[Schwarz et~al.(2020)Schwarz, Liao, Niemeyer, and Geiger]{schwarz2020graf}
Katja Schwarz, Yiyi Liao, Michael Niemeyer, and Andreas Geiger.
\newblock Graf: Generative radiance fields for 3d-aware image synthesis.
\newblock \emph{Advances in Neural Information Processing Systems}, 33:\penalty0 20154--20166, 2020.

\bibitem[Suvorov et~al.(2022)Suvorov, Logacheva, Mashikhin, Remizova, Ashukha, Silvestrov, Kong, Goka, Park, and Lempitsky]{suvorov2022resolution}
Roman Suvorov, Elizaveta Logacheva, Anton Mashikhin, Anastasia Remizova, Arsenii Ashukha, Aleksei Silvestrov, Naejin Kong, Harshith Goka, Kiwoong Park, and Victor Lempitsky.
\newblock Resolution-robust large mask inpainting with fourier convolutions.
\newblock In \emph{Proceedings of the IEEE/CVF winter conference on applications of computer vision}, pp.\  2149--2159, 2022.

\bibitem[Tang et~al.(2023)Tang, Ren, Zhou, Liu, and Zeng]{tang2023dreamgaussian}
Jiaxiang Tang, Jiawei Ren, Hang Zhou, Ziwei Liu, and Gang Zeng.
\newblock Dreamgaussian: Generative gaussian splatting for efficient 3d content creation.
\newblock \emph{arXiv preprint arXiv:2309.16653}, 2023.

\bibitem[Teed \& Deng(2020)Teed and Deng]{teed2020raft}
Zachary Teed and Jia Deng.
\newblock Raft: Recurrent all-pairs field transforms for optical flow.
\newblock In \emph{Computer Vision--ECCV 2020: 16th European Conference, Glasgow, UK, August 23--28, 2020, Proceedings, Part II 16}, pp.\  402--419. Springer, 2020.

\bibitem[Vachha \& Haque(2024)Vachha and Haque]{igs2gs}
Cyrus Vachha and Ayaan Haque.
\newblock Instruct-gs2gs: Editing 3d gaussian splats with instructions, 2024.
\newblock URL \url{https://instruct-gs2gs.github.io/}.

\bibitem[van~den Oord et~al.(2018)van~den Oord, Li, and Vinyals]{Oord2018RepresentationLW}
A{\"a}ron van~den Oord, Yazhe Li, and Oriol Vinyals.
\newblock Representation learning with contrastive predictive coding.
\newblock \emph{ArXiv}, abs/1807.03748, 2018.
\newblock URL \url{https://api.semanticscholar.org/CorpusID:49670925}.

\bibitem[Wang et~al.(2023{\natexlab{a}})Wang, Dutt, and Mitra]{Wang2023ProteusNeRFFL}
Binglun Wang, Niladri~Shekhar Dutt, and Niloy~Jyoti Mitra.
\newblock Proteusnerf: Fast lightweight nerf editing using 3d-aware image context.
\newblock \emph{ArXiv}, abs/2310.09965, 2023{\natexlab{a}}.
\newblock URL \url{https://api.semanticscholar.org/CorpusID:264146680}.

\bibitem[Wang et~al.(2022)Wang, Chai, He, Chen, and Liao]{wang2022clip}
Can Wang, Menglei Chai, Mingming He, Dongdong Chen, and Jing Liao.
\newblock Clip-nerf: Text-and-image driven manipulation of neural radiance fields.
\newblock In \emph{Proceedings of the IEEE/CVF Conference on Computer Vision and Pattern Recognition}, pp.\  3835--3844, 2022.

\bibitem[Wang et~al.(2023{\natexlab{b}})Wang, Zhang, Huang, Zhang, and Yan]{Wang2023SemanticIE}
Ruibo Wang, Song Zhang, Ping Huang, Donghai Zhang, and Wei Yan.
\newblock Semantic is enough: Only semantic information for nerf reconstruction.
\newblock \emph{2023 IEEE International Conference on Unmanned Systems (ICUS)}, pp.\  906--912, 2023{\natexlab{b}}.
\newblock URL \url{https://api.semanticscholar.org/CorpusID:265354446}.

\bibitem[Xie et~al.(2023)Xie, Zong, Qiu, Li, Feng, Yang, and Jiang]{xie2023physgaussian}
Tianyi Xie, Zeshun Zong, Yuxin Qiu, Xuan Li, Yutao Feng, Yin Yang, and Chenfanfu Jiang.
\newblock Physgaussian: Physics-integrated 3d gaussians for generative dynamics.
\newblock \emph{arXiv preprint arXiv:2311.12198}, 2023.

\bibitem[Xu et~al.(2022)Xu, Xu, Philip, Bi, Shu, Sunkavalli, and Neumann]{xu2022point}
Qiangeng Xu, Zexiang Xu, Julien Philip, Sai Bi, Zhixin Shu, Kalyan Sunkavalli, and Ulrich Neumann.
\newblock Point-nerf: Point-based neural radiance fields.
\newblock In \emph{Proceedings of the IEEE/CVF Conference on Computer Vision and Pattern Recognition}, pp.\  5438--5448, 2022.

\bibitem[Xu et~al.(2023)Xu, Yang, Mo, Pan, Yi, and Guibas]{Xu_2023_CVPR}
Xiaomeng Xu, Yanchao Yang, Kaichun Mo, Boxiao Pan, Li~Yi, and Leonidas Guibas.
\newblock Jacobinerf: Nerf shaping with mutual information gradients.
\newblock In \emph{Proceedings of the IEEE/CVF Conference on Computer Vision and Pattern Recognition (CVPR)}, pp.\  16498--16507, June 2023.

\bibitem[Yang et~al.(2023)Yang, Gao, Zhou, Jiao, Zhang, and Jin]{yang2023deformable}
Ziyi Yang, Xinyu Gao, Wen Zhou, Shaohui Jiao, Yuqing Zhang, and Xiaogang Jin.
\newblock Deformable 3d gaussians for high-fidelity monocular dynamic scene reconstruction.
\newblock \emph{arXiv preprint arXiv:2309.13101}, 2023.

\bibitem[Ye et~al.(2023)Ye, Danelljan, Yu, and Ke]{ye2023gaussian}
Mingqiao Ye, Martin Danelljan, Fisher Yu, and Lei Ke.
\newblock Gaussian grouping: Segment and edit anything in 3d scenes.
\newblock \emph{arXiv preprint arXiv:2312.00732}, 2023.

\bibitem[Yuan et~al.(2022)Yuan, Sun, Lai, Ma, Jia, and Gao]{Yuan2022NeRFEditingGE}
Yu-Jie Yuan, Yang-Tian Sun, Yu-Kun Lai, Yuewen Ma, Rongfei Jia, and Lin Gao.
\newblock Nerf-editing: Geometry editing of neural radiance fields.
\newblock \emph{2022 IEEE/CVF Conference on Computer Vision and Pattern Recognition (CVPR)}, pp.\  18332--18343, 2022.
\newblock URL \url{https://api.semanticscholar.org/CorpusID:248665648}.

\bibitem[Zhang et~al.(2023)Zhang, Kundu, Funkhouser, Guibas, Su, and Genova]{zhang2023nerflets}
Xiaoshuai Zhang, Abhijit Kundu, Thomas Funkhouser, Leonidas Guibas, Hao Su, and Kyle Genova.
\newblock Nerflets: Local radiance fields for efficient structure-aware 3d scene representation from 2d supervision.
\newblock In \emph{Proceedings of the IEEE/CVF Conference on Computer Vision and Pattern Recognition}, pp.\  8274--8284, 2023.

\bibitem[Zhi et~al.(2021)Zhi, Laidlow, Leutenegger, and Davison]{zhi2021place}
Shuaifeng Zhi, Tristan Laidlow, Stefan Leutenegger, and Andrew~J Davison.
\newblock In-place scene labelling and understanding with implicit scene representation.
\newblock In \emph{Proceedings of the IEEE/CVF International Conference on Computer Vision}, pp.\  15838--15847, 2021.

\bibitem[Zhou et~al.(2023)Zhou, Chang, Jiang, Fan, Zhu, Xu, Chari, You, Wang, and Kadambi]{zhou2023feature}
Shijie Zhou, Haoran Chang, Sicheng Jiang, Zhiwen Fan, Zehao Zhu, Dejia Xu, Pradyumna Chari, Suya You, Zhangyang Wang, and Achuta Kadambi.
\newblock Feature 3dgs: Supercharging 3d gaussian splatting to enable distilled feature fields.
\newblock \emph{arXiv preprint arXiv:2312.03203}, 2023.

\end{thebibliography}
\clearpage
\appendix
\section{Implementation Details}
\begin{figure}[!t]
    \includegraphics[width=\textwidth]{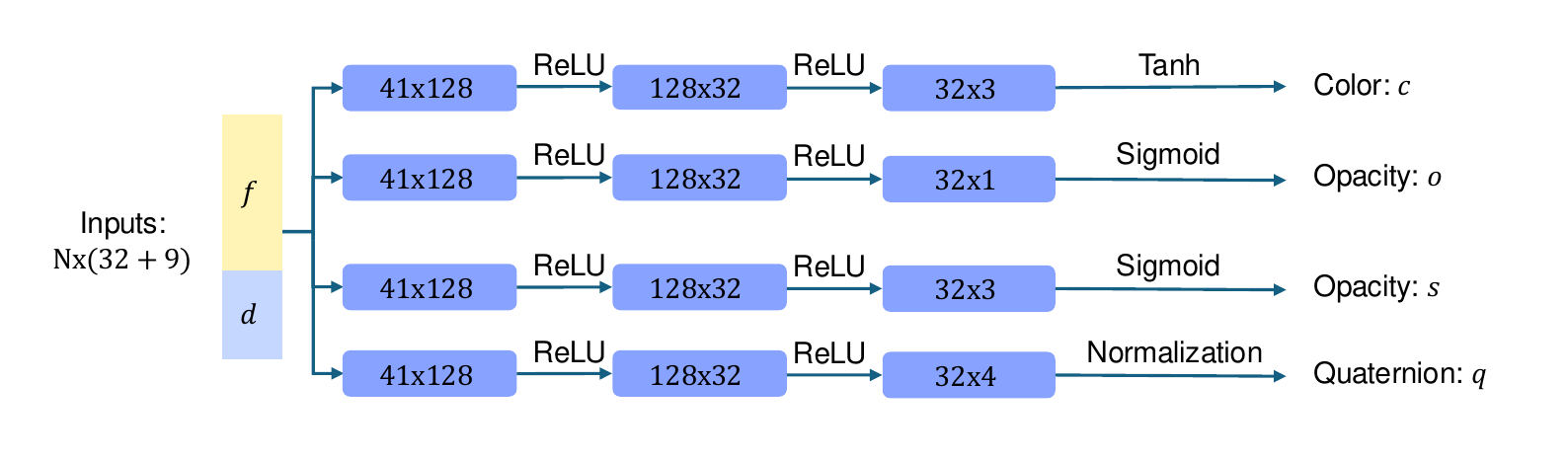}
    \vspace{-14pt}
    \caption{The structure of attribute decoding network. For each 3D Gaussian, we use different branches to predict attributes (opacity, color, scale, and quaternion). The inputs are features of Gaussians with view direction.}
    \label{fig:arc}
\end{figure}
\textbf{attribute decoding network.} The attribute decoding network is composed of four different branches, including the color branch $\mathbf{\Phi}_c$, opacity branch $\mathbf{\Phi}_o$, and branches for the covariance $\mathbf{\Phi}_s$ and $\mathbf{\Phi}_q$. They are all implemented by MLPs with RELU activation. The detailed structure is illustrated in Fig.~\ref{fig:arc}, where the output activation of different branches is inspired from~\citet{Lu2023ScaffoldGSS3}. The centroid $x$ of each Gaussian is initialized by the pre-computed point cloud from SfM points~\citep{ozyecsil2017survey} and separately optimized. The hidden dimension of each feature
is $32$ and initialized to be zero.

\textbf{Training hyperparamters.} During training, we set the hyperparameters $\lambda_{MI}=0.1$, $\lambda_{R}=0.1$ and $k=5$. We adopt the Adam optimizer~\cite{kingma2014adam} when shaping the attribute decoding network, with a learning rate of $0.01$. The gradient that we use to shape the motion network is $\partial h = \frac{\partial h^{(l)}}{\partial W^{(l)}}$ of each branch by default, where $l$ is the linear layer before the final output layer. We train the network for $1500$ iterations, sampling a random view and a batch of $512$ 3D Gaussians at each iteration to compute the training loss.

When training in the style of previous mutual information shaping in JacobiNeRF~\citep{Xu_2023_CVPR} to produce the results of JacobiGS, we modify the sampling batch size from $512$ to $48$ due to the large memory consumption. This is because the shaping method in JacobiNeRF requires the direct calculation of Jacobians $\partial \mathbf{\Phi}$, which necessitates the preservation of the entire computation graph.

\textbf{Training without using the tracker for mask association.} We use DEVA~\citep{cheng2023tracking} to associate the masks produced from SAM~\citep{kirillov2023segany} for training. However, the assumption of using the tracker to associate masks from different views is that the multi-view training images could be treated as a video. It will not hold when training views are sparse and random. In light of this, we introduce another shaping version without mask association, which is supervised by projecting the Jacobians to 2D for a sampled view:
\begin{align}
    \partial \mathbf{\Phi}(p) = \sum_{i\in\mathcal{N}}{\partial \mathbf{\Phi}_i\alpha_i\prod_{j=1}^{i-1}(1-\alpha_j)},
\end{align}
where $\partial \mathbf{\Phi}_i$ is the Jacobian of the $i$-th Gaussian 
and $\alpha_i$ is the blending weight that is calculated with the opacity, projected 2D covariance of the Gaussian, and the location of the pixel. By projecting the Jacobians of 3D Gaussians to 2D space, we can now sample Jacobians of pixels instead of 3D Gaussians for contrastive learning. We sample a random view and $512$ pixels of it for each iteration, supervised by the 2D mask of this view which is generated from SAM. We keep all other hyperparameters the same as the version of mask association. Note that we still use $\partial h$ to replace $\partial \mathbf{\Phi}$ for Jacobian shaping according to Eq.~\ref{eq:activation}.

\textbf{Speed-up module.} For the MI shaping version without mask association, it would be time-consuming when projecting Jacobians to the 2D plane as the dimension of the Jacobian is high. We then propose a speed-up module that modifies the rasterization process in the CUDA pipeline. As we only need to sample a small batch of Jacobians of pixels at each iteration, we add a mask during the CUDA forward and backward process to avoid the calculation of non-sampled pixels for Jacobians projection.

\section{Additional experimental results}
\subsection{scene editing}
\begin{figure}[!t]
    \includegraphics[width=\textwidth]{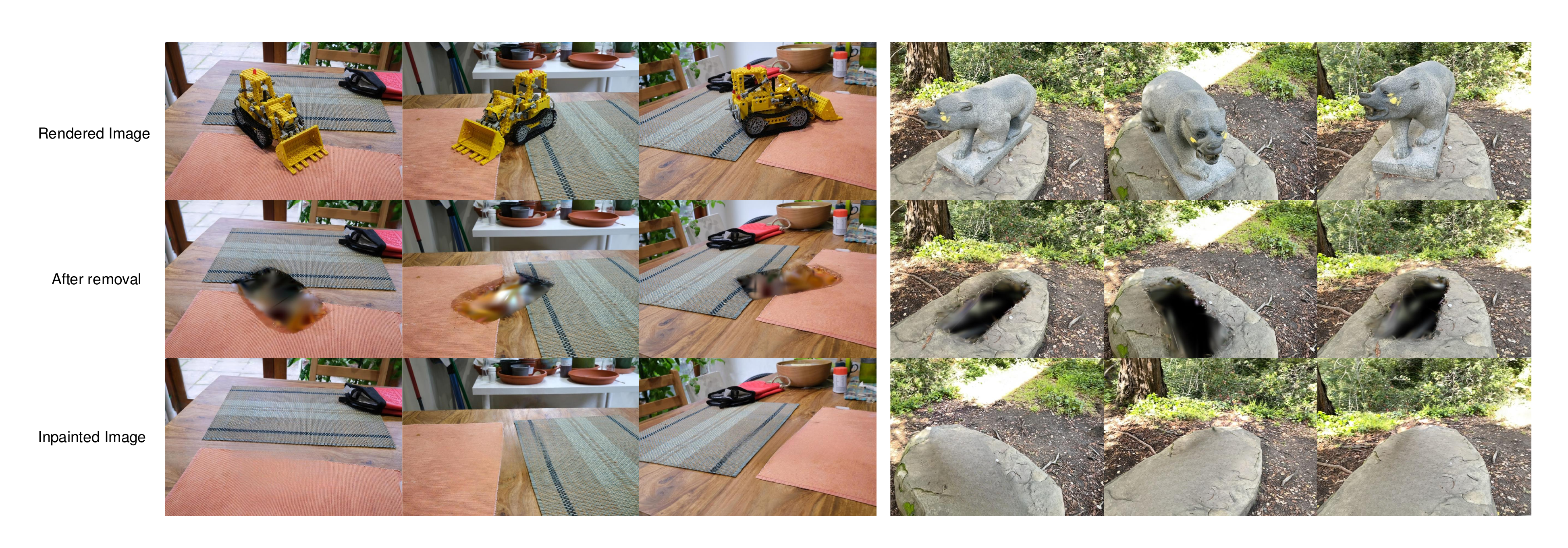}
    \vspace{-14pt}
    \caption{3D object Inpainting of our method. We use LAMA~\citep{suvorov2022resolution} to inpaint the images after object removal, and then finetune Gaussians based on the inpainted 2D images.}
    \label{fig:inpaint}
\end{figure}

\textbf{3D object inpainting.} Based on the 3D object removal, we could further accomplish the task of inpainting. Inspired by Gaussian Grouping~\citep{ye2023gaussian}, we first detect the noisy area after object removal through the mask association by DEVA~\citep{cheng2023tracking}, and then use LAMA~\citep{suvorov2022resolution} to inpaint the noisy deletion area for each view. Finally, we finetune Gaussians with similar training process according to the 2D inpainted images for about $5$ minutes to get the 3D inpainting results. Fig.~\ref{fig:inpaint} demonstrates the effectiveness of our method for 3D object inpainting.

\begin{figure}[!t]
    \includegraphics[width=\textwidth]{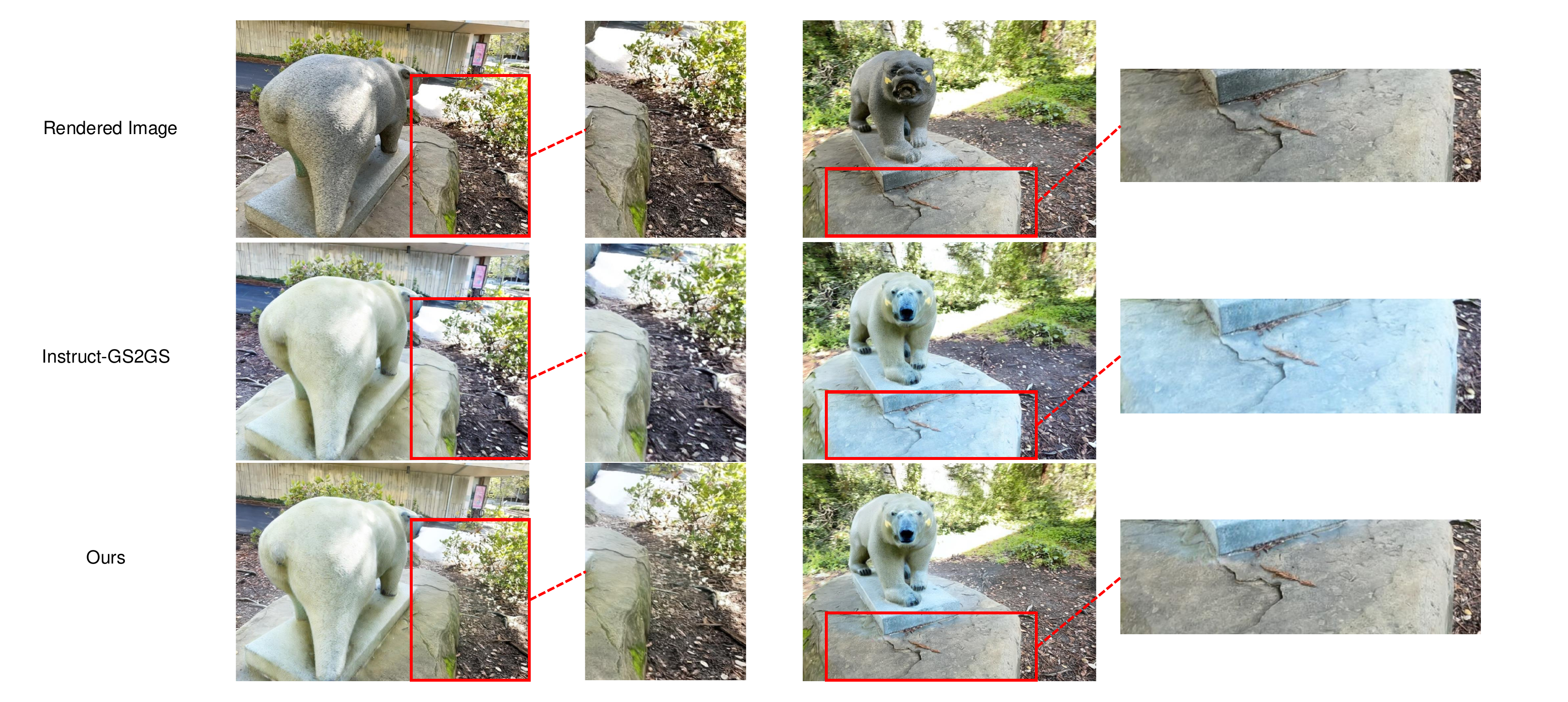}
    \vspace{-14pt}
    \caption{{3D object style transfer. We use Instruct-Pix2Pix~\citep{brooks2023instructpix2pix} to edit the 2D images (prompt: \textit{"Turn the bear into a polar bear"}), and then finetune Gaussians based on the edited 2D images. We compare our method with Instruct-GS2GS~\citep{igs2gs}, which is also a 3DGS-based style transfer method}.}
    \label{fig:style}
\end{figure}

\textbf{3D object style transfer.} {By selecting an object (similar to object removal), we could also achieve 3D object style transfer. First, we select all correlated Gaussians of the object through 3D segmentation. Second, we freeze the parameters of other Gaussians and start finetuning. During finetuning, we use the edited 2D ground truth images as supervision which are produced by Instruct-Pix2Pix~\citep{brooks2023instructpix2pix}, updating the original dataset. The pipeline is the same as Instruct-GS2GS. During finetuning, We employ L1 loss inside the 2D mask of the selected object and LPIPS loss within the bounding box that encloses the mask. From Fig.~\ref{fig:style}, we could see that our method significantly reduced the influence of the area that is not related to the targeted object. Instruct-GS2GS changes the hue of the whole figure and produce artifacts after editing.}

\begin{figure}[!t]
    \includegraphics[width=\textwidth]{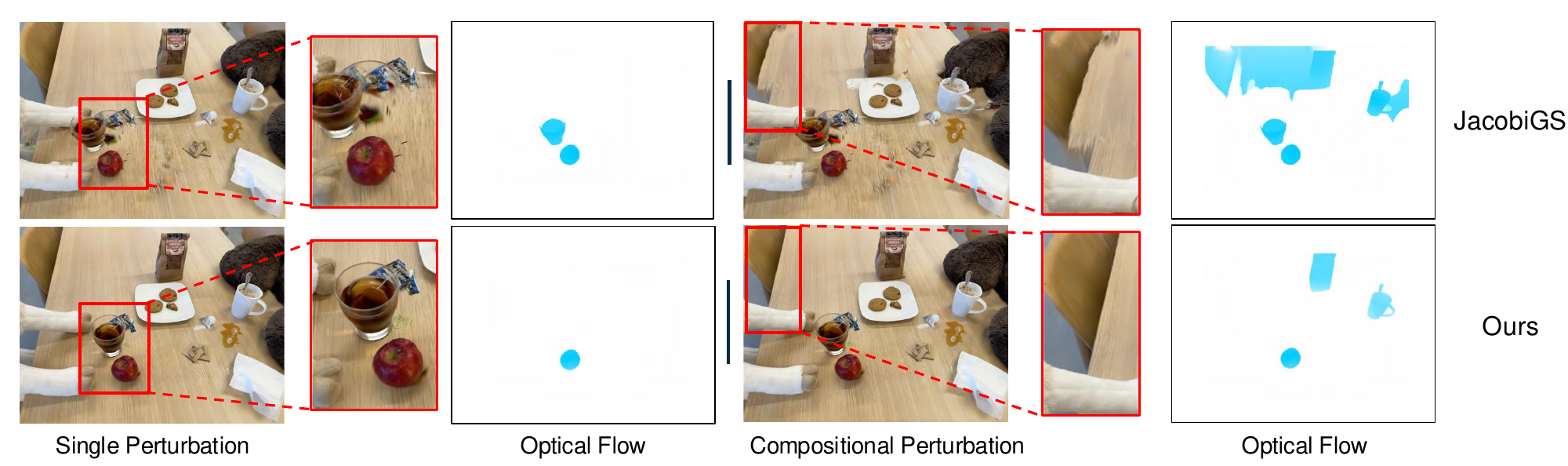}
    \vspace{-14pt}
    \caption{Visual comparison of multi-object editing between JacobiGS~\citep{Xu_2023_CVPR} and ours, where we slightly perturb three different objects: apple, cup, and bag. We use RAFT~\citep{teed2020raft} to estimate the optical flow for visualizing the movement.}
    \label{fig:compo}
\end{figure}

\textbf{Multi-object editing.} To achieve edit multiple objects in the same scene, we simply add Jacobians of selected Gaussians of different objects to the network parameters.
As shown in Fig.~\ref{fig:compo}, JacobiGS has difficulty maintaining scene integrity when combining Jacobians of two or more objects for perturbations. For instance, the table nearby is also affected by the perturbation (see the red box comparison of the right part in Fig.~\ref{fig:compo}). Conversely, our method effectively maintains robust correlation shaping, demonstrating superior performance in preserving the scene under multi-object editing. 

\subsection{Outdoor Scenarios}
\begin{figure}[!t]
    \includegraphics[width=\textwidth]{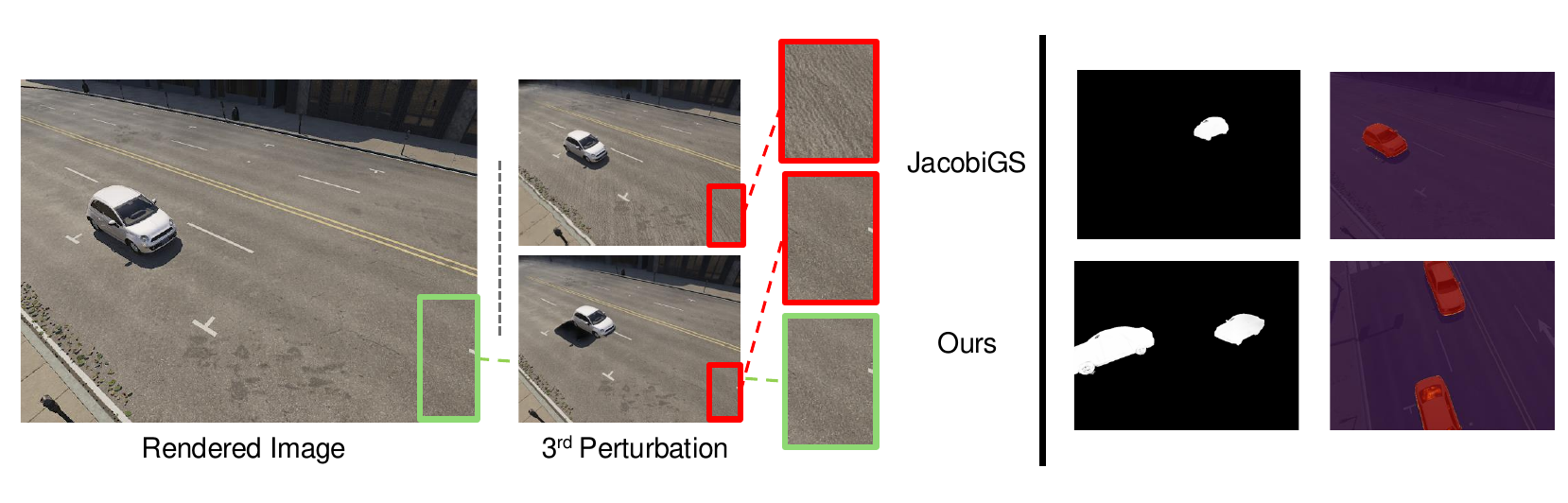}
    \vspace{-14pt}
    \caption{Qualitative results on NeRDS
360~\citep{irshad2023neo}. Left: Comparison of consecutive editions when moving the car. Right: The gallery of segmentation mask and relevance map from different views(when perturbing a single Gaussian), obtained by our method. }
    \label{fig:neo}
\end{figure}

To test the generalizability of our method, we employ our shaping method in NeRDS 360~\citep{irshad2023neo}, which is an outdoor unbounded dataset comprising 75 unbounded and diverse scenes. From Fig.~\ref{fig:neo}(a), we can clearly see that our method moves the car without creating any artifacts. JacobiGS fails to make the car move while changing the texture of the surrounding road. Fig.~\ref{fig:neo}(b) illustrated the ability to detect objects like cars in the 3D space, showing the potential of our method to be applied in practical scenarios.

\subsection{Ablation studies}
\begin{figure}[!t]
    \includegraphics[width=\textwidth]{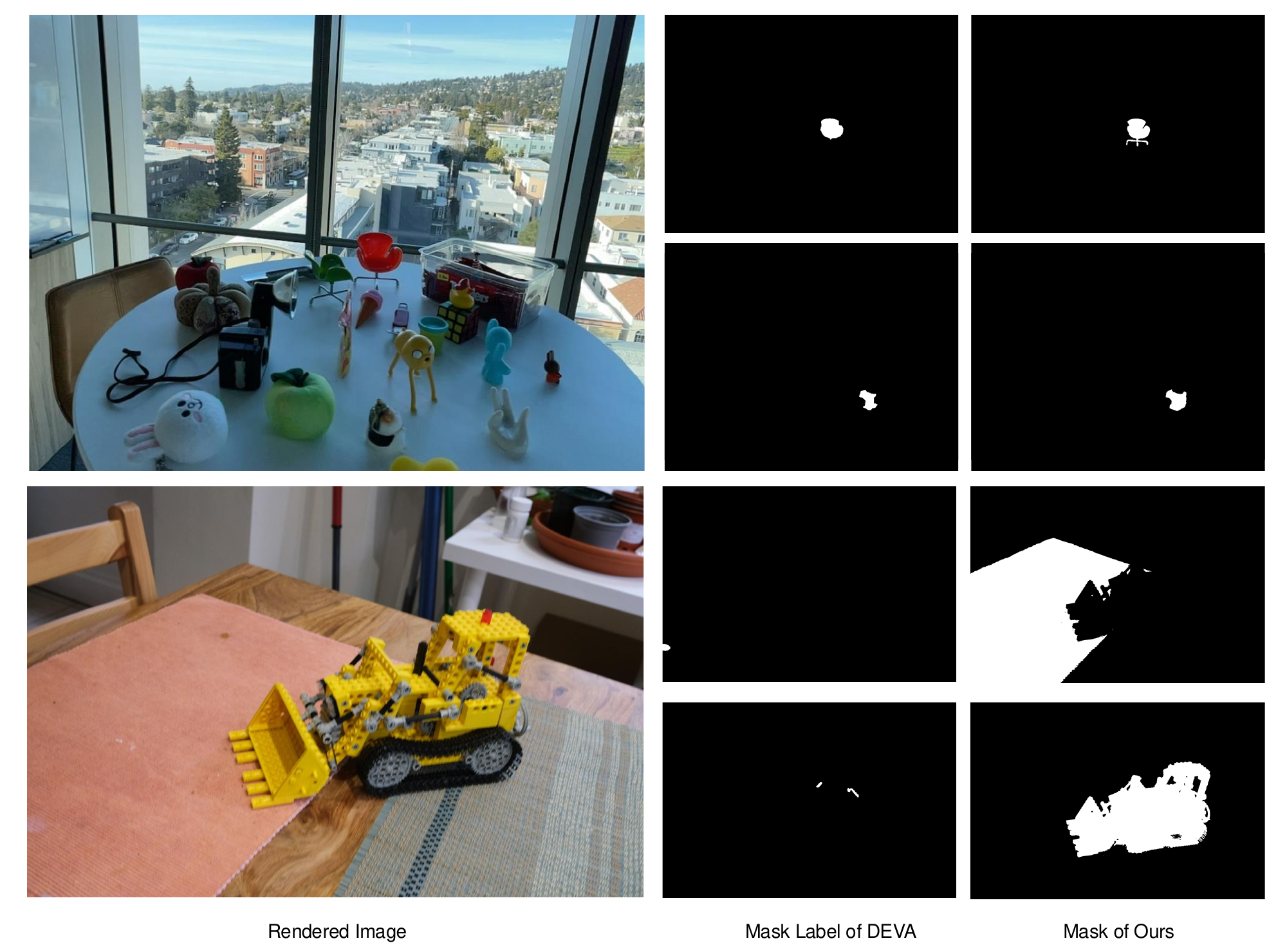}
    \vspace{-20pt}
    \caption{Robustness of our method to the multi-granularity and errors of masks. When masks from supervision contain errors (e.g., missing legs of the chair) with multi-granularity (e.g., a small part of the tractor is segmented in a certain view), our correlation shaping successfully addresses these problems and produces correct segmentation masks.}
    \label{fig:robust}
    \vspace{-6pt}
\end{figure}

\textbf{Robustness of correlation shaping.} Though the masks generated by SAM~\citep{kirillov2023segany} with DEVA~\citep{cheng2023tracking} are associated from different views, this strategy still suffers from the multi-granularity across different views and 2D segmentation errors. Fig.~\ref{fig:robust} demonstrates that our method resolves these problems by shaping the correlations between 3D Gaussians. Given a small part of the object as the prompt, our method can produce the segmentation of the entire object, showing that Gaussians inside the object are greatly associated.

\begin{figure}[!t]
    \includegraphics[width=\textwidth]{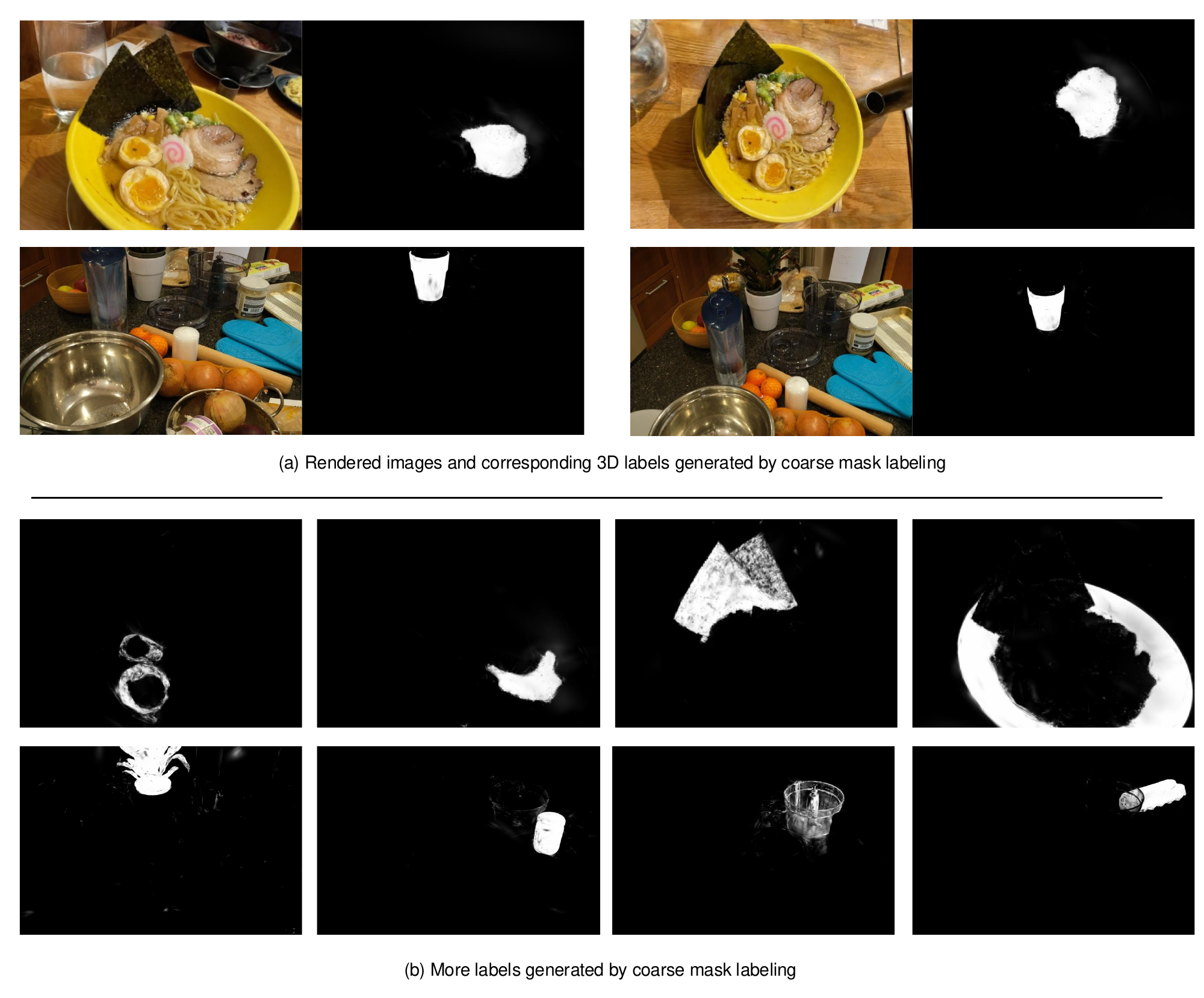}
    \vspace{-20pt}
    \caption{(a) Without any optimization, we produce 3D labels that are consistent across different views. (b) More examples of our 3D labels.done}
    \label{fig:coarse}
    \vspace{-6pt}
\end{figure}

\textbf{Quality of our 3D labels.} We visualize the quality of 3D labels in Fig.\ref{fig:coarse}, which are generated using the coarse mask labeling method from Eq.\ref{eq:mask}. As illustrated, these 3D labels are consistent across multiple views (Fig.\ref{fig:coarse}(a)) and exhibit high quality (Fig.\ref{fig:coarse}(b)) without requiring any optimization.

\begin{figure}[!t]
    \includegraphics[width=\textwidth]{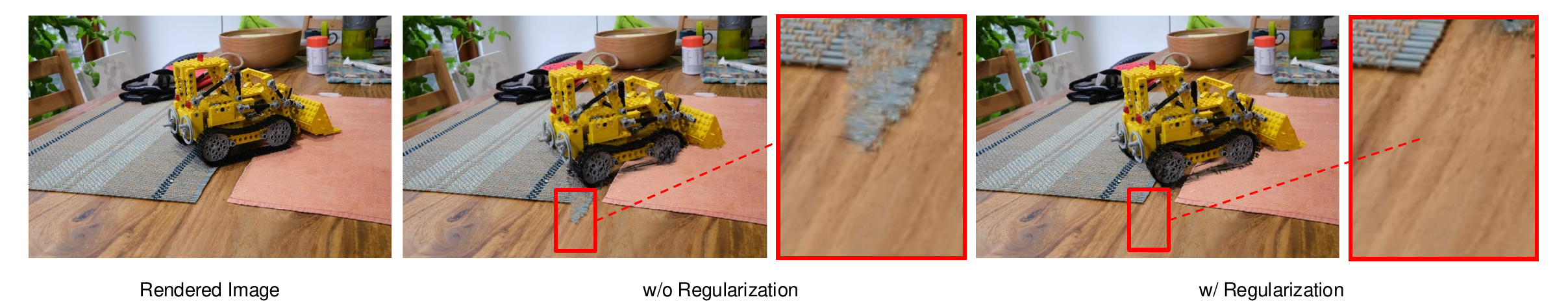}
    \vspace{-20pt}
    \caption{The impact of regularization loss when moving the lego tractor.}
    \label{fig:ab}
    \vspace{-6pt}
\end{figure}

\textbf{Regularization loss.} During finetuning the attribute decoding network, we introduce an extra regularization loss to help shape the Jacobians. Here Fig.~\ref{fig:ab} shows the ablation study of this regularization loss. From the area, the area in the red box apparently illustrates the impact of the regularization loss. When perturbing a Gaussian of the tractor, the one with regularization loss can maintains the minimal affects to other objects, while the one without it produces some artifacts to the surroundings. It demonstrates that the regularization loss can better shape the correlation structure of the tangent space of the motion network. 

% \textbf{Transfer to compression methods.}

\section{Proof of Derivation}
Given that the attribute decoding networks in neural fields (e.g., NeRFs and 3DGS) are MLPs, we focus on perturbing the weights $W^{(l)}$ of the $l$-th linear layer in the network $\mathbf{\Phi}_a$ (i.e., $\theta_D = W^{(l)}$), expressed as:
$h^{(l)}=W^{(l)}\sigma(h^{(l-1)})+b^{(l)}$. 
Let $\sigma$ denote the non-linear activation function, $b^{(l)}$ be the bias, and $h^{(l)}$ be the (hidden) output of the $l$-th layer.

To expose the relationship between the gradient of a Gaussian $\mathbf{g}_i$, $\partial \mathbf{\Phi}_i = \frac{\partial \mathbf{\Phi}(f(\mathbf{g}_i);\theta)}{\partial \theta_D}$, and the perturbed weight $W^{(l)}$, we consider the gradient of an arbitrary scalar output $z$ in the attribute decoding network with respect to $W^{(l)}$:
\begin{align}
    z = Y(h^{(l)})=Y(W^{(l)}\sigma(h^{(l-1)})+b^{(l)}),
\end{align}
where $Y$ denotes all the transformations after $l$-th layer, $z$ denotes an arbitrary scalar in the vector output of $\mathbf{\Phi}$. Then the differential $\mathrm{d}{z}$ could be written as:
\begin{equation}
\begin{aligned}
\mathrm{d}{z}&=\frac{\partial z}{\partial h^{(l)}}^\top\mathrm{d}{h^{(l)}}=\sigma(h^{(l-1)})\frac{\partial z}{\partial h^{(l)}}^\top\mathrm{d}{W^{(l)}}
=\text{tr}(\sigma(h^{(l-1)})\frac{\partial z}{\partial h^{(l)}}^\top\mathrm{d}{W^{(l)}}),
\end{aligned}
\end{equation}
as $\mathrm{d}{z}=\text{tr}(\frac{\partial z}{\partial W^{(l)}}^\top\mathrm{d}{W^{(l)}})$, we could get that:
\begin{align}
    \frac{\partial z}{\partial W^{(l)}}=\frac{\partial z}{\partial h^{(l)}}\sigma(h^{(l-1)})^\top,
\end{align}
which is a 2D matrix that has the same shape with the $W^{(l)}$. Then the cosine similarity of Jacobians $\partial \mathbf{\Phi}$ between two Gaussians are:
\begin{equation}
\begin{aligned}
\cos(\partial \mathbf{\Phi}_i,\partial{\mathbf{\Phi}}_j)&=\frac{(\frac{\partial z(\mathbf{x}_i)}{\partial h^{(l)}(\mathbf{x}_i)}\sigma(h^{(l-1)}(\mathbf{x}_i))^\top) \cdot (\frac{\partial z(\mathbf{x}_j)}{\partial h^{(l)}(\mathbf{x}_j)}\sigma(h^{(l-1)}(\mathbf{x}_j))^\top)}
{\lVert\partial\mathbf{\Phi}_i\rVert\lVert\partial\mathbf{\Phi}_j\rVert}\\
&=\frac{(\frac{\partial z(\mathbf{x}_i)}{\partial h^{(l)}(\mathbf{x}_i)} \cdot \frac{\partial z(\mathbf{x}_j)}{\partial h^{(l)}(\mathbf{x}_j)}) (\sigma(h^{(l-1)}(\mathbf{x}_i))^\top \cdot \sigma(h^{(l-1)}(\mathbf{x}_j))^\top)}
{\lVert\partial\mathbf{\Phi}_i\rVert\lVert\partial\mathbf{\Phi}_j\rVert},
\end{aligned}
\end{equation}
because $\partial h = \frac{\partial h^{(l)}}{\partial W^{(l)}}$ could be formulated as a 3D tensor, which could be conceptualized as a repeated collection of gradients and each gradient is the vector $\sigma(h^{(l-1)})$. If $\sigma(h^{(l-1)})$ is {\it well-shaped} satisfying object-level mutual relations (i.e., the absolute value of cosine similarity of $\sigma(h^{(l-1)})$ approaches 1 if $\mathbf{g}_i$ and $\mathbf{g}_j$ belong to the same object, or approaches 0 if not), then $\partial h$ should also satisfy. In this case, if $g_i$ and $g_j$ are affiliated to different objects:
\begin{equation}
\begin{aligned}
    \cos(\partial \mathbf{\Phi}_i,\partial{\mathbf{\Phi}}_j)&=\frac{(\frac{\partial z(\mathbf{x}_i)}{\partial h^{(l)}(\mathbf{x}_i)} \cdot \frac{\partial z(\mathbf{x}_j)}{\partial h^{(l)}(\mathbf{x}_j)}) (\sigma(h^{(l-1)}(\mathbf{x}_i))^\top \cdot \sigma(h^{(l-1)}(\mathbf{x}_j))^\top)}
    {\lVert\partial\mathbf{\Phi}_i\rVert\lVert\partial\mathbf{\Phi}_j\rVert}
    \\
    &=\frac{(\frac{\partial z(\mathbf{x}_i)}{\partial h^{(l)}(\mathbf{x}_i)} \cdot \frac{\partial z(\mathbf{x}_j)}{\partial h^{(l)}(\mathbf{x}_j)}) (0)}
    {\lVert\partial\mathbf{\Phi}_i\rVert\lVert\partial\mathbf{\Phi}_j\rVert}\\
    &=0,
\end{aligned}
\label{eq:neg_con}
\end{equation}
Eq.~\ref{eq:neg_con} indicates that if $\cos(\partial h_i, \partial h_j)=0$ when $g_i$ and $g_j$ are affiliated to different objects, then $\cos(\partial\mathbf{\Phi}_i,\partial{\mathbf{\Phi}}_j)=0$. Meanwhile, $\sigma(h^{(l-1)})$ would not change since the input of $\mathbf{\Phi}_a$ is the fixed $(f(\mathbf{g}_i), d)$. Thus $\cos(\partial h_i, \partial h_j)=0$ would be consistent during the perturbation. Therefore, we could have the following lemma:

\textbf{Lemma 1.} (\textit{Orthogonality}) After any number of perturbations, the orthogonality of Jacobians between two Gaussians $g_i \in \mathcal{G}_A$ 
and $g_j \in \mathcal{G}_B$ 
that are affiliated to different objects ($\mathcal{G}_A \neq \mathcal{G}_B$) would remain consistent:
\begin{equation}
    \cos (\partial{\mathbf{\Phi}^{(d)}_i},\partial{\mathbf{\Phi}^{(d)}_j}) = \cos (\partial{h^{(d)}_i},\partial{h^{(d)}_j}) = \cos (\partial{h^{(0)}_i},\partial{h^{(0)}_j}), d \in \mathbb{N}^+,
\end{equation}
In contrast, we have another lemma:

\textbf{Lemma 2.} (\textit{Similarity}) After any number of perturbations, the similarity of Jacobians between two Gaussians $g_i \in \mathcal{G}_A$ and $g_j \in \mathcal{G}_B$ that are affiliated to the same objects ($\mathcal{G}_A = \mathcal{G}_B$) would remain consistent:
\begin{equation}
    \cos (\partial{\mathbf{\Phi}^{(d)}_i},\partial{\mathbf{\Phi}^{(d)}_j}) \approx \cos (\partial{h^{(d)}_i},\partial{h^{(d)}_j}) = \cos (\partial{h^{(0)}_i},\partial{h^{(0)}_j}), d \in \mathbb{N}^+,
\end{equation}
\textit{Proof of Lemma 2}: Consider the set of Jacobians $J_A=\{\partial \mathbf{\Phi}_i, g_i \in \mathcal{G}_A\}$ of all Gaussians in an arbitrary object $\mathcal{G}_A$, they all would be orthogonal to the Jacobians of Gaussians from other objects, according to \textbf{Lemma 1} if $\partial h$ is {\it well-shaped}. Thus, the representation space of $J_A$ would be very narrow. If the representation space of $\partial \mathbf{\Phi}$ is compact enough (e.g., the dimension of the representation space of $\partial \mathbf{\Phi}$ is identical to the number of objects in the scene), then the similarity $\cos (\partial{\mathbf{\Phi}^{(d)}_i},\partial{\mathbf{\Phi}^{(d)}_j})=1$ when $\mathcal{G}_A=\mathcal{G}_B$. In this extreme case, the set of Jacobians from all the objects in the scene is a group of canonical basis~\citep{Bronson1969MatrixMA} of the representation space of $\partial \mathbf{\Phi}$.

Combining \textbf{Lemma 1} and \textbf{Lemma 2}, we could have the conclusion in Eq.~\ref{eq:trans}.

\end{document}